%% file: pra4.tex
\documentclass[11pt,twoside,english]{llncs}
\usepackage[T1]{fontenc}
\usepackage[latin1]{inputenc}
\usepackage{a4}
\usepackage{setspace}
\usepackage{amsmath}
\usepackage{graphicx}
\usepackage{amssymb}
\usepackage{times}
 \usepackage{gastex}
 \usepackage{pspictpg}

\makeatletter
\usepackage{graphicx}
\usepackage{epsfig}
\usepackage{epic}  
\usepackage{float}
\usepackage{color}

\usepackage[nice]{nicefrac}

\usepackage{dsfont}

\usepackage{babel}
\makeatother
\begin{document}


\newcommand{\marque}[2]{\marginpar{\begin{center}     \color{red} \fbox{{\LARGE \textbf{#1}}} \end{center}\fbox{\includegraphics[]{../../../../../Users/fdenis/Desktop/downloads/SIL-NotesMerise2004.pdf}

\begin{minipage}[c]{.94\marginparwidth}

#2 

\end{minipage}

}}}\newcommand{\TODO}[1]{\marque{TODO}{#1}}\newcommand{\TRAD}[1]{{\color{red} \textbf{#1}}\marque{Traduction}{#1}}

\newcommand{\ssi}{if\hbox{}f }\newcommand{\ie}{i.e.}\newcommand{\wrt}{with regard to }

\newcommand{\N}{\bbbn}

\newcommand{\Z}{\bbbz}

\newcommand{\Q}{\bbbq}

\newcommand{\IR}{\bbbr}

\newcommand{\C}{\bbbc}

\newcommand{\K}{\bbbk}

\newcommand{\card}[1]{\left|#1\right|}

\newcommand{\ens}[2]{\left\{  #1\vphantom{#2}\right|\left.\vphantom{#1}#2\right\}  }

\newcommand{\Res}[1]{\mathrm{Res}\left(#1\right)}

\newcommand{\res}[1]{\mathrm{res}\left(#1\right)}

\newcommand{\eps}{\varepsilon}

\newcommand{\conv}[1]{\mathrm{conv}\left(#1\right)}

\newcommand{\ensvide}{\varnothing}

\newcommand{\red}[2]{{\rm red}\left(#1,#2\right)}

\newcommand{\sat}[1]{{\rm sat}\left(#1\right)}

\newcommand{\redstar}[1]{\mathrm{red}^{*}\left(#1\right)}

\newcommand{\SL}[1]{\mathcal{S}\left(#1\right)}

\newcommand{\SA}[2]{\mathcal{S}_{\mathrm{#1}}}

\newcommand{\SPFA}{\SA{PA}{\Sigma}}

\newcommand{\SPRFA}{\SA{PRA}{\Sigma}}

\newcommand{\SPDFA}{\SA{PDA}{\Sigma}}

\newcommand{\cov}[2]{{\rm MaxCov}\left(#1,#2\right)}

\newcommand{\stcov}[2]{\mathrm{MaxCov}\left(#2,#1\right)}

\newcommand{\can}[1]{\mathrm{can}\left(#1\right)}

\newcommand{\nfrac}[2]{\nicefrac{#1}{#2}}

\newcommand{\pbar}{\overline{P}}

\newcommand{\Pbar}[1]{\overline{P}\left(#1\right)}

\newcommand{\Pbara}[2]{\overline{P_{#1}}\left(#2\right)}

\newcommand{\pbara}[1]{\overline{P_{#1}}}

\newcommand{\Qreach}{Q_{\mathrm{reach}}}

\title{Rational stochastic languages}

\author{Fran\c{c}ois Denis \and Yann Esposito }

\institute{LIF-CMI, UMR CNRS 6166\\
 39, rue F. Joliot Curie \\
 13453 Marseille Cedex 13 \\
 FRANCE\\
 Fax: 04 91 11 36 02\\
 \email{fdenis,esposito@cmi.univ-mrs.fr}}

\titlerunning{Rational stochastic languages}

\author{Fran\c{c}ois Denis \and Yann Esposito}

\authorrunning{Fran\c{c}ois Denis et al.}

\institute{LIF-CMI, UMR 6166\\
 39, rue F. Joliot Curie \\
 13453 Marseille Cedex 13 FRANCE\\
 \email{fdenis,esposito@cmi.univ-mrs.fr}}

\maketitle

\begin{abstract}The goal of the present paper is to provide a systematic
  and comprehensive study of \emph{rational stochastic languages} over
  a semiring $K\in \{{\mathbb Q}, {\mathbb Q}^+, {\mathbb R}, {\mathbb
    R}^+\}$. A rational stochastic language is a probability
  distribution over a free monoid $\Sigma^*$ which is rational over
  $K$, that is which can be generated by a multiplicity automata with
  parameters in $K$. We study the relations between the classes of
  rational stochastic languages ${\cal S}_{K}^{rat}(\Sigma)$. We
  define the notion of \emph{residual} of a stochastic language and we
  use it to investigate properties of several subclasses of rational
  stochastic languages. Lastly, we study the representation of
  rational stochastic languages by means of multiplicity automata.
\end{abstract}

\bibliographystyle{alpha}
\section{Introduction}
In probabilistic grammatical inference, data often arise in the form
of a finite sequence of words $w_1, \ldots, w_n$ over some predefined
alphabet $\Sigma$.  These words are assumed to be independently drawn
according to a fixed but unknown probability distribution over
$\Sigma^*$.  Probability distributions over free monoids $\Sigma^*$
are called \emph{stochastic languages}.  A usual goal in grammatical
inference is to try to infer an approximation of this distribution in
some class of probabilistic models, such as \emph{probabilistic
  automata}. A probabilistic automaton (PA) is composed of a
\emph{structure}, which is a finite automaton (NFA), and
\emph{parameters} associated with states and transitions, which
represent the probability for a state to be initial, terminal or the
probability for a transition to be chosen. It can easily be shown that
probabilistic automata have the same expressivity as Hidden Markov
Models (HMM), which are heavily used in statistical
inference~\cite{DupontDenisEsposito05}.  Given the structure $A$ of a
probabilistic automaton and a sequence of words $S$, computing
parameters for $A$ which maximize the likelihood of $S$ is NP-hard
\cite{AbeWarmuth92}. In practical cases however, algorithms based on
the E.M. (\emph{Expectation-Maximization}) method
\cite{DempsterLairdRubin77} can be used to compute approximate values.
On the other hand, inferring a probabilistic automaton (structure and
parameters) from a sequence of words is a widely open field of
research. Most results obtained so far only deal with restricted
subclasses of PA, such as Probabilistic Deterministic Automata (PDA),
i.e. probabilistic automata whose structure is deterministic (DFA) or
Probabilistic Residual Automata (PRA), i.e. probabilistic automata
whose structure is a residual finite state automaton
(RFSA)\cite{CarrascoOncina94,CarrascoOncina99,HigueraThollard00,EspositoLemayDenisDupont2002,DenisEsposito04}.

In other respects, it can be noticed that stochastic languages are
particular cases of \emph{formal power series} and that probabilistic
automata are also particular cases of \emph{multiplicity automata},
notions which have been extensively studied in the field of formal
language
theory\cite{SalomaaSoittola78,BerstelReutenauer84,Sakarovitch03}.
Therefore, stochastic languages which can be generated by multiplicity
automata are special cases of \emph{rational languages}. We call them
\emph{rational stochastic languages}. The goal of the present paper is
to provide a systematic and comprehensive study of \emph{rational
  stochastic languages} so as to bring out properties that could be
useful for a grammatical inference purpose. Indeed, considering the
objects to infer as special cases of rational languages makes it
possible to use the powerful theoretical tools that have been
developed in that field and hence, give answers to many questions that
naturally arise when working with them: is it possible to decide
within polynomial time whether two probabilistic automata generate the
same stochastic language? does allowing negative coefficients in
probabilistic automata extend the class of generated stochastic
languages? can a rational stochastic language which takes all its
values in ${\mathbb Q}$ always be generated by a multiplicity automata
with coefficients in ${\mathbb Q}$?  and so forth. Also, studying
\emph{rational stochastic languages} for themselves, considered as
objects of language theory, helps to bring out notions and properties
which are important in a grammatical inference pespective: for
example, we show that the notion of residual language (or derivative),
so important for grammatical
inference~\cite{DenisLemayTerlutte2002b,DenisLemayTerlutte04}, has a
natural counterpart for stochastic languages~\cite{DenisEsposito2003},
which can be used to express many properties of classes of stochastic
languages.

\emph{Formal power series }take their values in a semiring $K$: let us denote
by $K\langle\langle\Sigma\rangle\rangle$ the set of all formal power
series. Here, we only consider semirings ${\mathbb Q}$, ${\mathbb R}$,
${\mathbb Q}^+$ and ${\mathbb R}^+$. For any such semiring $K$, we
define the set ${\cal S}_K^{rat}(\Sigma)$ of rational stochastic
languages as the set of stochastic languages over $\Sigma$ which are
rational languages over $K$. For any two distinct semirings $K$ and
$K'$, the corresponding sets of rational stochastic languages are
distinct. We show that ${\mathbb R}$ is a Fatou extension of ${\mathbb
Q}$ for stochastic languages, which means that any rational stochastic
language over ${\mathbb R}$ which takes its values in ${\mathbb Q}$ is
also rational over ${\mathbb Q}$. However, ${\mathbb R}^+$ is not a
Fatou extension of ${\mathbb Q}^+$ for stochastic languages: there
exists a rational stochastic language over ${\mathbb R}^+$ which takes
its values in ${\mathbb Q}^+$ and which is not rational over ${\mathbb
Q}^+$.

For any stochastic language $p$ over $\Sigma$ and any word $u$ such
that $p(u\Sigma^*)\neq 0$, let us define the residual language
$u^{-1}p$ of $p$ with respect to $u$ by $u^{-1}p(w)=p(uw)/p(u\Sigma^*)$:
residual languages clearly are stochastic languages. We show that the
residual languages of a rational stochastic language $p$ over $K$ are
also rational over $K$. The residual subsemimodule $[Res(p)]$ of
$K\langle\langle\Sigma\rangle\rangle$ spanned by the residual
languages of any stochastic language $p$ may be used to express the
rationality of $p$: $p$ is rational iff $[Res(p)]$ is included in a
finitely generated subsemimodule of
$K\langle\langle\Sigma\rangle\rangle$. But when $K$ is positive, i.e.
$K={\mathbb Q}^+$ or $K={\mathbb R}^+$, it may happen that $[Res(p)]$
itself is not finitely generated.  We study the properties of two
subclasses of ${\cal S}_K^{rat}(\Sigma)$: the set ${\cal
  S}_K^{fingen}(\Sigma)$ composed of rational stochastic languages
over $K$ whose residual subsemimodule is finitely generated and the
set ${\cal S}_K^{fin}(\Sigma)$ composed of rational stochastic
languages over $K$ which have finitely many residual languages. We
show that for any of these two classes, ${\mathbb R}^+$ is a Fatou
extension of ${\mathbb Q}^+$: any stochastic language of ${\cal
  S}_{{\mathbb R}^+}^{fingen}(\Sigma)$ (resp. of ${\cal S}_{{\mathbb
    R}^+}^{fin}(\Sigma)$) which takes its values in ${\mathbb Q}^+$ is
an element of ${\cal S}_{{\mathbb Q}^+}^{fingen}(\Sigma)$ (resp. of
${\cal S}_{{\mathbb Q}^+}^{fin}(\Sigma)$). We also show that for any
element $p$ of ${\cal S}_K^{fingen}(\Sigma)$, there exists a unique
minimal subset of residual languages of $p$ which generates
$[Res(p)]$.

Then, we study the representation of rational stochastic languages by
means of multiplicity automata. We first show that the set of
multiplicity automata with parameters in ${\mathbb Q}$ which generate
stochastic languages is not recursive. Moreover, it contains no
recursively enumerable subset capable to generate the whole set of
rational stochastic languages over ${\mathbb Q}$. A stochastic
language $p$ is a formal series which has two properties: (i)
$p(w)\geq 0$ for any word $w$, (ii) $\sum_wp(w)=1$. We show that the
undecidability comes from the first requirement, since the second one
can be decided within polynomial time. We show that the set of
stochastic languages which can be generated by probabilistic automata
with parameters in ${\mathbb Q}^+$ (resp.${\mathbb R}^+$) exactly
coincides with ${\cal S}_{{\mathbb Q}^+}^{rat}(\Sigma)$ (resp. ${\cal
  S}_{{\mathbb R}^+}^{rat}(\Sigma)$). A probabilistic automaton $A$ is
called a Probabilistic Residual Automaton (PRA) if the stochastic
languages associated with its states are residual languages of the
stochastic languages $p_A$ generated by $A$. We show that the set of
stochastic languages that can be generated by probabilistic residual
automata with parameters in ${\mathbb Q}^+$ (resp.${\mathbb R}^+$)
exactly coincides with ${\cal S}_{{\mathbb Q}^+}^{fingen}(\Sigma)$
(resp. ${\cal S}_{{\mathbb R}^+}^{fingen}(\Sigma)$). We do not know
whether the class of PRA is decidable. However, we describe two
decidable subclasses of PRA capable of generating ${\cal
  S}_{K}^{fingen}(\Sigma)$ when $K={\mathbb Q}^+$ or $K={\mathbb
  R}^+$: the class of $K$-reduced PRA and the class of prefixial PRA.
The first one provides minimal representation in the class of PRA but
we show that the membership problem is PSPACE-complete.  The second
one produces more cumbersome representation but the membership problem
is polynomial. Finally, we show that the set of stochastic languages
that can be generated by probabilistic deterministic automata with
parameters in ${\mathbb Q}^+$ (resp.${\mathbb R}^+$) exactly coincides
with ${\cal S}_{{\mathbb Q}^+}^{fin}(\Sigma)$, which is also equal to
${\cal S}_{{\mathbb Q}}^{fin}(\Sigma)$ (resp. ${\cal S}_{{\mathbb
    R}^+}^{fin}(\Sigma)$, which is also equal to ${\cal S}_{{\mathbb
    R}}^{fin}(\Sigma)$).

We recall some properties on rational series, stochastic languages and
multiplicity automata in Section~\ref{prelim}. We define and study
rational stochastic languages in Section~\ref{rsl}. The relations
between the classes of rational stochastic languages are studied in
Subsection~\ref{crsl}. Properties of the residual languages of
rational stochastic languages are studied in Subsection~\ref{rlrsl}. A
characterisation of rational stochastic languages in terms of stable
subsemimodule is given in Subsection~\ref{charrsl}. Classes ${\cal
S}_{K}^{fingen}(\Sigma)$ and ${\cal S}_{K}^{fin}(\Sigma)$ are defined
and studied in Subsection~\ref{fingenfin}. The representation of
rational stochastic languages by means of multiplicity automata is
given in Section~\ref{marsl}.

\section{Preliminaries}\label{prelim}
\subsection{Rational series}
In this section, we recall some definitions and results on rational
series. For more information, we invite the reader to consult
\cite{SalomaaSoittola78,BerstelReutenauer84,Sakarovitch03}.

Let $\Sigma$ be a finite \emph{alphabet}, and $\Sigma^{*}$ be the set
of words on $\Sigma$. The empty word is denoted by $\varepsilon$ and
the length of a word $u$ is denoted by $|u|$.  The number of
occurrences of the letter $x$ in the word $w$ is denoted by
$|w|_x$. For any integer $k$, we denote by $\Sigma^k$ the set $\{u\in \Sigma^{*}\ |\ |u|=k\}$ and by $\Sigma^{\leq k}$ the set $\{u\in \Sigma^{*}\ | \ |u|\leq k\}$. We denote by $<$ the length-lexicographic order on
$\Sigma^*$. For any word $u\in \Sigma^*$ and any language $L\subseteq
\Sigma^*$, let $uL=\{uv\in \Sigma^*| v\in L\}$ and $u^{-1}L=\{v\in
\Sigma^*| uv\in L\}$. A subset $P$ of $\Sigma^{*}$ is \emph{prefixial}
if for any $u,v\in \Sigma^*$, $uv\in P\Rightarrow u\in P$.

A \emph{semiring} is a set $K$ with two binary operations $+$ and
$\cdot$ and two constant elements 0 and 1 such that
\begin{enumerate}
\item $\langle K,+,0\rangle$ is a commutative monoid,
\item $\langle K,\cdot,1\rangle$ is a monoid,
\item the distribution laws  $a\cdot(b+c)=a\cdot b + a\cdot c$ and $(a+b)\cdot c=a\cdot c + b\cdot c$ hold,
\item $0\cdot a=a\cdot 0=0$ for every $a$.
\end{enumerate}

A semiring is \emph{positive} if the sum of two elements different from
0 is different from 0.

The semirings we consider here are the field of rational numbers
${\mathbb Q}$, the field of real numbers ${\mathbb R}$, ${\mathbb
Q}^+$ and ${\mathbb R}^+$, respectively the non negative elements of
${\mathbb Q}$ and ${\mathbb R}$; ${\mathbb Q}^+$ and ${\mathbb R}^+$
are positive semirings.

Let $\Sigma$ be a finite alphabet and $K$ a semiring. A \emph{formal power series}
is a mapping $r$ of $\Sigma^*$ into $K$. 
The values $r(w)$ where $w\in \Sigma^*$ are
referred to as the \emph{coefficients} of
the series, and $r$ is written as a formal sum $r=\sum_{w\in
\Sigma^*}r(w)w$.
The set of all formal power series is denoted by $K\langle\langle
\Sigma\rangle\rangle$. Given $r$, the subset of $\Sigma^*$ defined by
$\{w|r(w)\neq 0\}$ is the \emph{support} of $r$ and denoted by
$supp(r)$. A \emph{polynomial} is a series whose support is
finite. The subset of $K\langle\langle \Sigma\rangle\rangle$ consisting of
all polynomials is denoted by $K\langle \Sigma\rangle$.

We denote by 0 the series all of whose coefficients equal 0. We denote
by 1 the series whose coefficient for $\varepsilon$ equals 1, the
remaining coefficients being equal to 0. The \emph{sum} of two series
$r$ and $r'$ in $K\langle\langle \Sigma\rangle\rangle$ is defined by
$r+r'=\sum_{w\in \Sigma^*}(r(w)+r'(w))w.$ The multiplication of a
series $r$ by a scalar $a\in K$ is defined by $ar=\sum_{w\in
\Sigma^*}a\cdot r(w)w.$ The Cauchy product of two series $r$ and $r'$ is
defined by $rr'=\sum_{w\in
\Sigma^*}\left(\sum_{w_1w_2=w}r(w_1)\cdot r'(w_2)\right)w.$ These operations furnish $K\langle\langle \Sigma\rangle\rangle$ with the structure of a semiring with $K\langle\Sigma \rangle$ as a subsemiring. The Hadamard product of two series $r$ and $r'$ is defined by $r \odot  r'=\sum_{w\in
\Sigma^*} r(w)r'(w)w.$ 

A series $r$ is \emph{quasiregular} if $r(\epsilon)=0$. Quasiregular
series have the property that for every $w\in \Sigma^*$, there exist
finitely many integers $i$ such that $r^i(w)\neq 0$ where the exponent $i$ of 
$r^i$ refers to the Cauchy product. Let $r$ be a quasiregular series,
$r^*$ (resp. $r^+$) is defined by $r^*(w)=\sum_{i\geq 0}r^i(w)$ (resp.
$r^+(w)=\sum_{i\geq 1}r^i(w)$).

A subsemiring $R$ of $K\langle\langle \Sigma\rangle\rangle$ is
\emph{rationally closed} if $r^+\in R$ for every quasiregular element
$r$ of $R$. The family $K^{rat}\langle\langle \Sigma\rangle\rangle$ of
$K$-rational series over $\Sigma$ is the smallest rationally closed
subset of $K\langle\langle \Sigma\rangle\rangle$ which contains all
polynomials. When $K$ is commutative, the Hadamard product of two
rational series is a rational series.

Let $K$ be a semiring and let $m,n$ be two integers. Let us denote by
$K^{m\times n}$ the set of $m\times n$ matrices whose elements belong
to $K$ and by $I_m$ the matrix whose diagonal elements are equal to 1
and whose all other elements are null. Note that $K^{m\times m}$
forms a semiring.

A series $r$ is \emph{recognizable} if there exists a multiplicative
homomorphism $\mu: \Sigma^* \rightarrow K^{n\times n}, n\geq 1$, and
two matrices $\lambda\in K^{1\times n}, \gamma\in K^{n\times 1}$ such
that for every $w\in \Sigma^*$, $r(w)=\lambda \mu(w) \gamma$. The
tuple $(\lambda, \mu, \gamma)$ is called an $n$ dimensional
\emph{linear representation} of $r$. A linear representation of $r$ is
said to be \emph{reduced} if its dimension is minimal.

 Let us denote by
$K^{rec}\langle\langle \Sigma\rangle\rangle$ the set of all
recognizable series.

\begin{theorem}\label{recrat} \cite{Schutzenberger61}
The families $K^{rat}\langle\langle \Sigma\rangle\rangle$ and
$K^{rec}\langle\langle \Sigma\rangle\rangle$ coincide.
\end{theorem}

Let $K$ be a semiring. Then a commutative monoid V is called a
\emph{$K$-semimodule} if there is an operation $\cdot$ from $K\times
V$ into $V$ such that for any $a,b\in K, v,w\in V$,
\begin{enumerate}
\item $(ab)\cdot v = a\cdot(b\cdot v)$,
\item $(a+b)\cdot v = a\cdot v + b\cdot v$ and  $a\cdot(v+w)=a\cdot v + a\cdot w$,
\item $1\cdot v=v$ and $0\cdot v =0$.
\end{enumerate}
If $S$ is a subset of a $K$-semimodule $V$, the subsemimodule
$[S]$ generated by $S$ is the smallest of all subsemimodules of $V$
containing $S$. It can be proved that $[S]=\{a_1s_1+\ldots + a_ns_n
| n\in {\mathbb N}^*, a_i\in K, s_i\in S\}.$


Let us consider the semimodule $K^{\Sigma^*}$ of all functions $F:
\Sigma^*\rightarrow K$. For any word $u$ of $\Sigma^*$ and any
function $F$ of $K^{\Sigma^*}$, we define a new function $\dot{u}F$ by
$\dot{u}F(v)=F(uv)$ for any word $v$. The operator transforming $F$ into
$\dot{u}F$ is linear: for any $F, G\in K^{\Sigma^*}$ and
$a\in K$, $\dot{u}\left(a\cdot F\right)=a\cdot \dot{u}F$ and
$\dot{u}(F+G)=\dot{u}F+\dot{u}G$. A subset $B$ of
$K^{\Sigma^*}$ is called \emph{stable} if the conditions $u\in
\Sigma^*$ and $F\in B$ imply that $\dot{u}F\in B$.

\begin{theorem}\label{caract}\cite{Fliess74,Jacob75}
Suppose that $K$ is a commutative semiring and $r$ belongs to $K\langle\langle \Sigma\rangle\rangle$. Then the following three conditions are equivalent:
\begin{enumerate}
\item $r$ belongs to $K^{rat}\langle\langle \Sigma\rangle\rangle$;
\item the subsemimodule of $K\langle\langle \Sigma\rangle\rangle$ generated
by $\{\dot{u}r|u\in \Sigma^*\}$ is contained in a finitely generated stable
subsemimodule of $K^{\Sigma^*}$;
\item $r$ belongs to a finitely generated stable subsemimodule of
$K^{\Sigma^*}$.
\end{enumerate}
\end{theorem}

When $K$ is not a field, it may happen that a series $r$
belongs to a finitely generated stable subsemimodule of
$K\langle\langle \Sigma\rangle\rangle$, and hence is a rational
series, while the stable subsemimodule generated by $\{\dot{u}r|u\in
\Sigma^*\}$ is not finitely generated. An example of this situation
will be provided on Example~\ref{ex:nonfingen}.

Two linear representations $(\lambda, \mu, \gamma)$ and $(\lambda',
\mu', \gamma')$ of a rational series $r$ are \emph{similar} if there
exists an inversible matrix $m\in K^{n\times n}$ such that
$\lambda'=\lambda m, \mu' w = m^{-1}\mu w m$ for any word $w$ and $\gamma' =
m^{-1}\gamma$.

\begin{theorem}\label{similar}\cite{Schutzenberger61,Fliess74} 
Assume that $K$ is a commutative field. Then any two reduced linear
representations $(\lambda, \mu, \gamma)$ and $(\lambda', \mu',
\gamma')$ of a rational series $r$ are similar. The dimension of any reduced linear representation of $r$ is also  the dimension of the vector subspace generated by $\{\dot{u}r|u\in \Sigma^*\}$.
\end{theorem}

Let $K$ be a subsemiring of $K'$. $K'$ is said to be a \emph{Fatou extension} of
$K$ if every rational series over $K'$ with coefficients in $K$ is a
rational series over $K$. It has been shown in~\cite{Fliess74} that
when $K$ and $K'$ are commutative fields then $K'$ is a Fatou extension
of $K$. Therefore, ${\mathbb R}$ is a Fatou
extension of ${\mathbb Q}$: any rational series over
${\mathbb R}$ which only takes rational values is a rational
series over ${\mathbb Q}$: 
${\mathbb R}^{rat}\langle\langle
  \Sigma\rangle\rangle \cap {\mathbb Q}\langle\langle
  \Sigma\rangle\rangle = {\mathbb Q}^{rat}\langle\langle
  \Sigma\rangle\rangle.$ It has also been proved that ${\mathbb
R^+}$ is not a Fatou extension of ${\mathbb
Q^+}$: ${\mathbb
Q^+}^{rat}\langle\langle \Sigma\rangle\rangle \subsetneq {\mathbb
R^+}^{rat}\langle\langle \Sigma\rangle\rangle\cap {\mathbb
Q^+}\langle\langle \Sigma\rangle\rangle$.

\subsection{Stochastic languages}\label{SL}

A \emph{stochastic language} is a formal series $p$ which takes its
values in ${\mathbb R}^+$ and such that $\sum_{w\in
\Sigma^*}p(w)=1$. For any stochastic language $p$ and any language
$L\subseteq \Sigma^*$, the sum $\sum_{w\in L}p(w)$ is defined without
ambiguity. So, let us denote $\sum_{w\in L}p(w)$ by $p(L)$. The set of
all stochastic languages over $\Sigma$ is denoted by ${\cal
S}(\Sigma)$.  For any stochastic language $p$ and any word $u$ such
that $p(u\Sigma^*)\neq 0$, we define the stochastic
language $u^{-1}p$ by $$u^{-1}p(w)=\frac{p(uw)}{p(u\Sigma^*)}\cdot$$ $u^{-1}p$ is called the \emph{residual
language} of $p$ wrt $u$. Let us denote by $res(p)$ the set $\{u\in
\Sigma^*|\sum_{w\in \Sigma^*}p(uw)\neq 0\}$ and by $Res(p)$ the set
$\{u^{-1}p|u\in res(p)\}$. For any $K\in \{{\mathbb R}, {\mathbb R^+},
{\mathbb Q}, {\mathbb Q^+}\}$, define ${\cal
S}_K^{rat}(\Sigma)=K^{rat}\langle\langle\Sigma\rangle\rangle\cap {\cal
S}(\Sigma)$, the set of rational stochastic languages over $K$. Let
$S=\{s_1, \ldots, s_n\}$ be a finite subset of ${\cal S}(\Sigma)$. The
convex hull of $S$ in $K\langle\langle \Sigma\rangle\rangle$ is
defined by $conv_K(S)=\{s\in K\langle\langle \Sigma\rangle\rangle |
s=\alpha_1\cdot s_1 + \ldots + \alpha_n\cdot s_n$ where each
$\alpha_i\in K, \alpha_i\geq 0$ and $\alpha_1+ \ldots +
\alpha_n=1\}$. Clearly, any element of $conv_K(S)$ is a stochastic
language.

\begin{example}\label{ex:nonfingen}
Let $\Sigma=\{a\}$, and let $p_1$, $p_2$ and $p$ be the rational
stochastic languages over ${\mathbb R}^+$ defined on $\Sigma^*$ by
$$p_1(a^n)=2^{-(n+1)}, p_2(a^n)=3\cdot 2^{-(2n+2)} \textrm{ and
}p=(p_1+p_2)/2.$$

Check that
$$\dot{\overline{a^n}}p_1=\frac{p_1}{2^n}, \dot{\overline{a^n}}p_2=\frac{p_2}{2^{2n}}\textrm{ and }\dot{\overline{a^n}}p=\frac{2^np_1+p_2}{2^{2n+1}}$$
and
$$(a^n)^{-1}p_1=p_1, (a^n)^{-1}p_2=p_2\textrm{ and }(a^n)^{-1}p=\frac{2^np_1+p_2}{2^n+1}\cdot$$
Let ${\cal V}$ be the vector subspace of ${\mathbb R}\langle\langle
\Sigma\rangle\rangle$ generated by $p_1$ and $p_2$: ${\cal V}$ is
represented on Figure~\ref{fig:exemple}. 
\begin{figure}[htbp]\label{fig:exemple}
  \input{nonfingen.pstex_t}
  \caption{The stable subsemimodule of ${\mathbb R}^+\langle\langle \Sigma\rangle\rangle$ generated by $p$ is equal to ${\cal V}_p$: it does not contains the halfline $]Op_1)$ and it is not finitely generated. }
  
\end{figure}
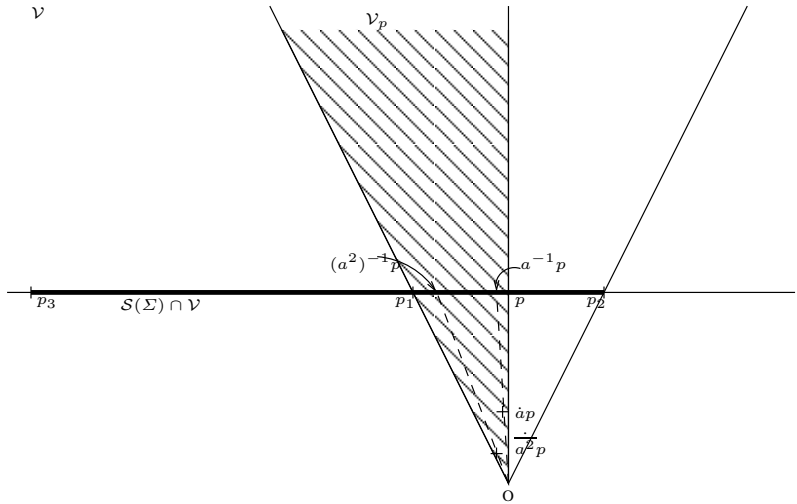
The subsemimodule of
${\mathbb R}^+\langle\langle \Sigma\rangle\rangle$ generated by $p_1$
and $p_2$ corresponds to the closed halfcone ${\cal C}$ delimited by
the halflines $[Op_1)$ and $[Op_2)$.  The line $(p_1p_2)$ is composed
of the rational series $r$ in ${\cal V}$ which satisfy $\sum_{w\in
\Sigma^*}r(w)=1$. Let $q=\alpha p_1+(1-\alpha) p_2$. The constraint
$q(a^n)\geq 0$ is equivalent to the inequality
$$(2^{n+1}-3)\alpha +3\geq 0.$$ The series $q$ such that $q(a^n)\geq
0$ for any integer $n$ must satisfy $$0\leq \alpha\leq 3.$$ Let $p_3=3p_1-2p_2$. The stochastic
languages in ${\cal V}$ are the points of the line $(p_2p_3)$ which
lie between $p_2$ and $p_3$.

Let ${\cal V}_p$ be the subsemimodule of ${\mathbb R}^+\langle\langle
\Sigma\rangle\rangle$ generated by $\{\dot{u}p|u\in \Sigma^*\}$. 
Check that ${\cal V}_p=\{t(\alpha p_1+(1-\alpha) p_2) | 1/2\leq \alpha
<1, t \in {\mathbb R}^+\}$ 
and that ${\cal V}_p$ is not finitely generated.
\end{example}

\subsection{Automata}\label{MA}

A \textit{non deterministic finite automaton} (NFA) is a tuple
$\langle\Sigma,Q,Q_I,Q_T,\delta\rangle$ where $Q$ is a finite set of
states, $Q_I\subseteq Q$ is the set of initial states, $Q_T\subseteq
Q$ is the set of final states, $\delta$ is the \textit{transition
function} defined from $Q\times\Sigma$ to $2^{Q}$. Let $\delta$ also
denote the extended transition function defined from
$2^{Q}\times \Sigma^*$ to $2^{Q}$ by
$\delta(q,\varepsilon)=\{q\}$, $\delta(q,wx)=\cup_{q'\in
\delta(q,w)}\delta(q',x)$ and $\delta(R,w)=\cup_{q\in R}\delta(q,w)$ for any $q\in Q$, $R\subseteq Q$, $x\in \Sigma$ and $w\in
\Sigma^*$. An NFA is \textit{deterministic}
(DFA) if $Q_I$ contains only one element $q_{0}$ and if $\forall q\in
Q$, $\forall x\in\Sigma$, $\card{\delta(q,x)}\leq 1$. 


Let $K$ be a semiring. A $K$-\emph{multiplicity automaton (MA)}
  is a 5-tuple $\left\langle \Sigma,Q,\varphi,\iota,\tau\right\rangle
  $ where $Q$ is a finite set of states, $\varphi:Q\times\Sigma\times
  Q\rightarrow K$ is the transition function, $\iota:Q\rightarrow K$
  is the initialization function and $\tau:Q\rightarrow K$ is the
  termination function. Let $Q_I=\{q\in Q|\iota(q)\neq 0\}$ be the set
  of \emph{initial states} and $Q_T=\{q\in Q|\tau(q)\neq 0\}$ be the
  set of \emph{terminal states}. The \emph{support} of an MA
  $\left\langle \Sigma,Q,\varphi,\iota,\tau\right\rangle$ is the NFA
  $\langle \Sigma,Q,Q_I, Q_T, \delta\rangle$ where
  $\delta(q,x)=\{q'\in Q|\varphi(q,x,q')\neq 0\}$.  We extend the
  transition function $\varphi$ to $Q\times\Sigma^*\times Q$ by
  $\varphi(q,wx,r)=\sum_{s\in Q}\varphi(q,w,s)\varphi(s,x,r)$ and
  $\varphi(q,\varepsilon,r)=1$ if $q=r$ and $0$ otherwise, for any
  $q,r\in Q$, $x\in \Sigma$ and $w\in \Sigma^*$. For any finite subset
  $L\subset \Sigma^*$ and any $R\subseteq Q$, define
  $\varphi(q,L,R)=\sum_{w\in L, r\in R}\varphi(q,w,r)$. 

For any MA $A=\left\langle \Sigma,Q,\varphi,\iota,\tau\right\rangle $,
we define the series $r_A$ by $$r_A(w)=\sum_{q, r\in
Q}\iota(q)\varphi(q,w,r)\tau(r).$$ For any $q\in Q$, we define the
series $r_{A,q}$ by $r_{A,q}(w)=\sum_{r\in Q}\varphi(q,w,r)\tau(r).$

If the semiring $K$ is positive, it can be shown that the support of
the series $r_A$ defined by a $K$-multiplicity automaton is equal
to the language defined by the support of $A$. In particular,
$supp(r_A)$ is a regular language. This property is false in general when $K$ is
not positive.

Two MA $A$ and $A'$ are
\emph{equivalent} if they define the same series, i.e. if
$r_A=r_{A'}$.

Let $A=\left\langle \Sigma,Q,\varphi,\iota,\tau\right\rangle $ be a
$K$-MA and let $q\in Q$. Suppose that there exist coefficients
 $\alpha_{q'}\in K$ for $q'\in Q'=Q\setminus\{q\}$ such
that
$r_{A,q}=\sum_{q'\in Q'}\alpha_{q'} r_{A,q'}.$
Let $A'= \left\langle \Sigma,Q',\varphi',\iota',\tau'\right\rangle $
where 
\begin{itemize}
\item $\varphi'(r,x,s)=\varphi(r,x,s)+\alpha_s\varphi(r,x,q)$ for any $r,s\in Q'$ and $x\in \Sigma$,
\item $\iota'(r)=\iota(r)+\alpha_r\iota(q)$ for any $r\in Q'$,
\item $\tau'(r)=\tau(r)$ for any $r\in Q'$.
\end{itemize}

The multiplicity automaton $A'$ is called a $K$-\emph{reduction} of $A$. 
A multiplicity automaton $A$ is called $K$-\emph{reduced} if it has no
$K$-reduction.

 \begin{proposition}
 Let $A=\left\langle \Sigma,Q,\varphi,\iota,\tau\right\rangle $ be a $K$-MA and let $A'= \left\langle \Sigma,Q',\varphi',\iota',\tau'\right\rangle$ be a $K$-reduction of
 $A$. Then, for any state $q'\in Q'$,
 $r_{A',q'}=r_{A,q'}$. As a consequence, $r_{A'}=r_{A}$.
 \end{proposition}

 \begin{proof}
 Let $Q'=Q\setminus \{q\}$ and let $\alpha_{q'}\in K$ for any $q'\in Q'$
such that $r_{A,q}=\sum_{q'\in Q'}\alpha_{q'} r_{A,q'}.$  For any
state $r\in Q'$, we have $$r_{A',r}(\varepsilon )=\tau' (r)=\tau
(r)=r_{A,r}(\varepsilon ).$$ Now, assume that for any word $w$ of
length $\leq k$ and any state $r\in Q'$ we have
$r_{A',r}(w)=r_{A,r}(w)$. Let $x$ be a letter, we have:
 \begin{eqnarray*}
 r_{A',r}(xw) & = & \sum _{s\in Q'}\varphi'(r,x,s)r_{A',s}(w)=\sum _{s\in Q'}\left(\varphi (r,x,s)+\alpha_{s}\varphi (r,x,q)\right)r_{A,s}(w)\\
  & = & \sum _{s\in Q'}\varphi (r,x,s)r_{A,s}(w)+\varphi (r,x,q)\sum _{s\in Q'}\alpha_{s}r_{A,s}(w)\\
  & = & \sum _{s\in Q'}\varphi (r,x,s)r_{A,s}(w)+\varphi (r,x,q)r_{A,q}(w)\\
  & = & \sum _{s\in Q}\varphi (r,x,s)r_{A,s}(w)=r_{A,r}(xw).\\
 \end{eqnarray*}
  Hence, $r_{A',r}=r_{A,r}$ for any $r$ of $Q'$. Moreover,
 \begin{eqnarray*}
 r_{A'} & = & \sum _{s\in Q'}\iota'(s)r_{A,s}=\sum _{s\in Q'}\left(\iota (s)+\alpha_{s}\iota (q)\right)r_{A,s}\\\
  & = & \sum _{s\in Q'}\iota (s)r_{A,s}+\iota (q)\sum _{s\in Q'}\alpha_{s}r_{A,s}=\sum _{s\in Q}\iota (s)r_{A,s}= r_{A}.\\
 \end{eqnarray*} \qed
 \end{proof}

A state $q\in Q$ is \emph{accessible} (resp. \emph{co-accessible}) if there exists $q_0\in Q_I$ (resp. $q_t\in Q_T$) and
  $u\in \Sigma^*$ such that $\varphi(q_0,u,q)\neq 0$ (resp. $\varphi(q,u,q_t)\neq 0$).
An MA is \emph{trimmed} if all its states are accessible and
co-accessible. Given an MA $A$, a trimmed MA equivalent to $A$ can
efficiently be computed from $A$.


From now, we only consider trimmed MA.

We shall consider several subclasses of multiplicity automata, defined as follows:

A \emph{semi Probabilistic Automaton (semi-PA)} is an MA $\left\langle
\Sigma,Q,\varphi,\iota,\tau\right\rangle $ such that $\iota, \varphi$
and $\tau$ take their values in $[0,1]$, such that $\sum_{q\in
Q}\iota(q)\leq 1$ and for any state $q$,
$\tau(q)+\varphi(q,\Sigma,Q)\leq 1$.  Semi-PA generate rational series
over ${\mathbb R}^+$.

A \emph{Probabilistic Automaton (PA)} is a trimmed semi-PA $\left\langle
\Sigma,Q,\varphi,\iota,\tau\right\rangle $ such that
$\sum_{q\in Q}\iota(q)=1$ and for any state $q$,
$\tau(q)+\varphi(q,\Sigma,Q)=1$. Probabilistic automata generate stochastic languages.

\begin{proposition}\label{PASto}
Let $A=\left\langle \Sigma,Q,\varphi,\iota,\tau\right\rangle $ be a
$K$-semi-PA (resp. a $K$-PA). For $q\in Q$, $\sum_{w\in
\Sigma^*}r_{A,q}(w)\leq 1$ (resp. $\sum_{w\in \Sigma^*}r_{A,q}(w)=
1$). As a consequence, $\sum_{w\in
\Sigma^*}r_{A}(w)\leq 1$ (resp. $\sum_{w\in \Sigma^*}r_{A}(w)=
1$).
\end{proposition}
\begin{proof}
For any integer $k$ and any $q\in Q$, we have 
\begin{align*}
&\sum_{|w|\leq k+1}r_{A,q}(w)+\varphi(q,\Sigma^{k+2},Q)\\
&=\sum_{|w|\leq
k}r_{A,q}(w)+\sum_{r\in Q}\varphi(q,\Sigma^{k+1},r)\tau(r)+\sum_{r\in
Q}\varphi(q,\Sigma^{k+1},r)\varphi(r,\Sigma,Q)\\
&=\sum_{|w|\leq
k}r_{A,q}(w)+\sum_{r\in
Q}\varphi(q,\Sigma^{k+1},r)[\tau(r)+\varphi(r,\Sigma,Q)].
\end{align*}
 From this
relation, it is easy to infer by induction on $k$ that $$\sum_{|w|\leq
k}r_{A,q}(w)+\sum_{r\in Q}\varphi(q,\Sigma^{k+1},r)\leq 1 \textrm{
(resp. } =1)$$
 when $A$ is a semi-PA (resp. a PA).

A first consequence is that $$\sum_{w\in \Sigma^*}r_{A,q}(w)\leq
1\textrm{ and }\sum_{w\in \Sigma^*}r_{A}(w)=\sum_{w\in \Sigma^*}\sum_{q\in
Q}\iota(q)r_{A,q}(w)\leq 1.$$ Let $n=|Q|$. Since
$A$ is trimmed, there exists a word $u\in \Sigma^{\leq n-1}$ such that
$r_{A,q}(u)>0$. Therefore, there exists $\alpha<1$ such that
$\varphi(q,\Sigma^n,Q)<\alpha$. It can easily be shown, by induction
on the integer $k$, that $\varphi(q,\Sigma^{kn},Q)<\alpha^k$.

Now, when $A$ is a PA, we have  $$\sum_{w\in \Sigma^*}r_{A,q}(w)\geq \sum_{|w|<kn}r_{A,q}(w)=1-\varphi(q,\Sigma^{kn},Q)>1-\alpha^k$$
for any integer $
k$. Therefore, $$\sum_{w\in \Sigma^*}r_{A,q}(w)=1.$$
Finally, $$\sum_{w\in \Sigma^*}r_{A}(w)=\sum_{w\in \Sigma^*}\sum_{q\in
Q}\iota(q)r_{A,q}(w)=\sum_{q\in
Q}\iota(q)=1.$$\qed
\end{proof}
It can easily be deduced from Proposition~\ref{PASto} that a ${\mathbb
R}^+$-reduction of a PA is still a PA (the property is false in
general for a semi-PA).

 A \emph{Probabilistic Residual Automaton (PRA)} is a PA $\left\langle
\Sigma,Q,\varphi,\iota,\tau\right\rangle $ such that for any $q\in Q$,
there exists a word $u$ such that $r_{A,q}=u^{-1}r_A$. Check that a
${\mathbb R^+}$-reduction of a PRA is still a PRA, since the series
associated with the states remain unchanged within a reduction.

 A \emph{Probabilistic Deterministic Automaton (PDA)} is a PA whose
support is deterministic. Check that a PDA is a PRA. Therefore, a
${\mathbb R^+}$-reduction of a PDA is a PRA, but since reduction introduces
non-determinism, it is no longer a PDA.

\begin{figure}[h]
    \begin{center}
       \begin{picture}(120,50)
\put(5,35){$A$}
\letstate i0=(10,35)     
\letstate q0=(20,35) 
\drawtrans(i0,q0){1}
\drawstate(q0){$q_0$}     
\letstate q1=(40,35)      
\letstate t0=(50,35)     
\drawstate(q1){$q_1$}
\drawtrans(q0,q1){$a,0.5$}
\drawtrans(q1,t0){1}
\drawloop[t](q0){$b,0.5$} 
\put(60,35){$B$} 
\letstate i1=(65,35)     
\letstate q2=(75,35) 
\drawtrans(i1,q2){1}
\drawstate(q2){$q_0$}     
\letstate q3=(95,35)     
\drawstate(q3){$q_1$}
\drawtrans(q2,q3){$a,0.4;b,0.4$}
\drawloop[t](q2){$b,0.2$}      
\letstate t1=(105,35) 
\drawtrans(q3,t1){1} 
\put(5,15){$C$}
\letstate i2=(10,15)     
\letstate q4=(20,15) 
\drawtrans(i2,q4){0.5}
\drawstate(q4){$q_0$}     
\letstate q5=(40,15)      
\letstate i3=(50,15)     
\drawstate(q5){$q_1$}     
\letstate t3=(47,22) 
\drawtrans(q5,t3){0.4} 
\drawloop[t](q4){$b,0.2$} 
\drawcurvedtrans(q4,q5){$a,0.4;b,0.4$}
\drawcurvedtrans(q5,q4){$b,0.5$}
\drawtrans(i3,q5){0.5}
\put(60,15){$D$}
\letstate i4=(65,15)     
\letstate q6=(75,15) 
\drawtrans(i4,q6){0.5}
\drawstate(q6){$q_0$}     
\letstate q7=(95,15)      
\letstate i5=(105,15)     
\drawstate(q7){$q_1$}     
\letstate t4=(102,22) 
\drawtrans(q7,t4){0.4} 
\drawloop[t](q6){$b,0.2$} 
\drawcurvedtrans(q6,q7){$a,0.4;b,0.4$}
\drawcurvedtrans(q7,q6){$a,0.2;b,0.3$}
\drawtrans(i5,q7){0.5}

\end{picture}
\caption{Let us precise notations on automaton $A$: $q_0$ is the
unique initial state and $\iota(q_0)=1$, $q_1$ is the unique terminal
state and $\tau(q_1)=1$, $\varphi(q_0,a,q_1)=0.5$, $\varphi(q_0,b,q_0)=0.5$ and any other transitions satisfy $\varphi(q,x,q')=0$. $A$ is a PDA; $B$
is a PRA since $r_{B,q_0}=r_B$ and $r_{B,q_1}=a^{-1}r_B$; $C$ is also a PRA since
$r_{C_,q_0}=ab^{-1}r_C$ and
$r_{C_,q_1}=a^{-1}r_C$; it can easily be shown that $D$ is not a PRA. }
\label{MAs}
\end{center}
  \end{figure}
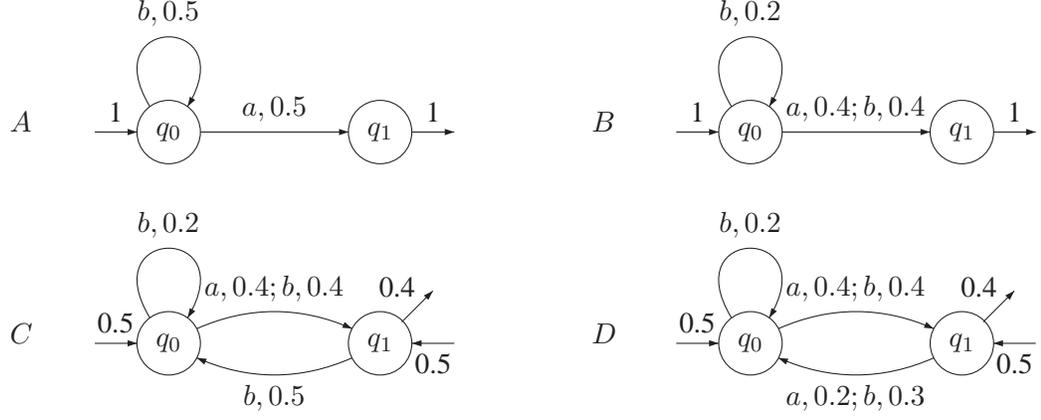

For any class $C$ of $K$-multiplicity automata, let us denote by ${\cal S}_K^C(\Sigma)$ the class of all stochastic languages which are recognized by an element of $C$.

\subsection{Equivalent representations of rational series}

  Stable finitely generated subsemimodules, linear representations and
  multiplicity automata provide us with several representations
  of rational series. The following classical claims show that they
  are equivalent: in particular, a series $r$ over $K$ is rational iff
  there exists a $K$-multiplicity automaton $A$ such that
  $r=r_A$. Moreover, any one of these representations can efficiently be derived from
  any other one. 

\begin{itemize}
\item[Claim 1] Let $M$ be a stable subsemimodule of $K\langle\langle
\Sigma\rangle\rangle$ generated by $r_1, \ldots, r_n$ and containing
the series $r$. Let $\alpha_i$ and $\alpha_{i,j}^x$ be coefficients in
$K$ defined for any letter $x$ and any $1\leq i,j\leq n$ such that
$$r=\sum_{i=1}^n\alpha_ir_i\textrm{ and
}\dot{x}r_i=\sum_{j=1}^n\alpha_{i,j}^xr_j.$$ Let $(\lambda, \mu,
\gamma)$ be the linear representation defined by
$\lambda[1,i]=\alpha_i$, $\mu(x)[i,j]=\alpha_{i,j}^x$ and
$\gamma[i,1]=r_i(\varepsilon)$ for any $1\leq i,j\leq n$ and any $x\in \Sigma$. Then $(\lambda, \mu,
\gamma)$ is a linear representation of $r$.
\item[Claim 2] Let $(\lambda, \mu, \gamma)$ be an $n$-dimensional linear
representation of $r$ and let $A=\left\langle
\Sigma,Q,\varphi,\iota,\tau\right\rangle $ be the MA defined by
$Q=\{1, \ldots, n\}$, $\iota(i)=\lambda[1,i]$, $\tau(i)=\gamma[i,1]$
and $\varphi(i,x,j)=\mu(x)[i,j]$. Then $r=r_A$.
\item[Claim 3] Let $A=\left\langle
    \Sigma,Q,\varphi,\iota,\tau\right\rangle $ be an MA and let $M$ be
  the subsemimodule generated by $\{r_{A,q}|q\in Q\}$. Then $M$ is a stable subsemimodule of $K\langle\langle
  \Sigma\rangle\rangle$ which contains $r_A$.
\end{itemize}

The proofs of these claims are classical. We give them for sake of completness.

\begin{proof}[Claim 1]
Let us prove by induction on the length of the word $w$ that for any word $w$, $\mu(w)\gamma=(r_1(w), \ldots, r_n(w))^t$. 
From definition, $\mu(\varepsilon)\gamma=\gamma=(r_1(\varepsilon), \ldots, r_n(\varepsilon))^t$. 

Suppose that the relation is proved for all words of length $\leq n$ and let $w\in \Sigma^n$ and $x\in \Sigma$. 

\begin{align*}
\mu(xw)\gamma&=\mu(x)\mu(w)\gamma\\
&=\mu(x)(r_1(w), \ldots, r_n(w))^t \textrm{ by induction hypothesis}\\
&=\left(\sum_{j=1}^n\alpha_{1,j}^xr_j(w), \ldots, \sum_{j=1}^n\alpha_{n,j}^xr_j(w)\right)^t\\
&=\left(\dot{x}r_1(w),\ldots, \dot{x}r_n(w)\right)^t\\
&=\left(r_1(xw), \ldots, r_n(xw)\right)^t.\\
\end{align*}

Now, for any word $w$, $$\lambda\mu(w)\gamma=\lambda(r_1(w), \ldots, r_n(w))^t=\sum_{i=1}^n\alpha_ir_i(w)=r(w).$$\qed

\end{proof}

\begin{proof}[Claim2]
For any word $w$, we have $$r_A(w)=\sum_{i,j=1}^n\iota(i)\varphi(i,w,j)\tau(j)=\sum_{i,j=1}^n=\lambda[1,i]\mu(w)[i,j]\gamma[i,1]=\lambda\mu(w)\gamma.$$\qed
\end{proof}

\begin{proof}[Claim3]
First note that $r_A=\sum_{q\in Q}\iota(q)r_{A,q}$ and therefore, $r_A\in M$.

Next, for any letter $x$, any word $w$ and any state $q\in Q$,
$$\dot{x}r_{A,q}(w)=r_{A,q}(xw)=\sum_{q'\in
Q}\varphi(q,x,q')r_{A,q'}(w)$$ and therefore,
$$\dot{x}r_{A,q}=\sum_{q'\in Q}\varphi(q,x,q')r_{A,q'}.$$ $M$ is a
stable subsemimodule of $K\langle\langle \Sigma\rangle\rangle$.  \qed
\end{proof}


These equivalent characterizations make it possible to transfer
definitions from one representation mode to another: check that an
$n$-dimensional linear representation of a rational series over $K$ is reduced
if and only iff the corresponding multiplicity automaton is
$K$-reduced.
  Also, results obtained using one representation can immediatly be
transfered to the other ones.

\subsection{Computing equivalence and reduction of MA} 

Deciding whether two NFA are equivalent is a PSPACE-complete
problem. However, deciding whether two MA are equivalent can be
achieved within polynomial time.

 \begin{proposition}\label{Prop:eqMA}
  It is decidable within polynomial time whether two MAs over ${\mathbb R}$ are equivalent.
\end{proposition}
\begin{proof}
Let $A$ and $A'$ be two MA and let $(\lambda, \mu,
\gamma)$  (resp. $(\lambda', \mu', \gamma')$) be an $n$-dimensional (resp.
$n'$-dimensional) linear representation of the rational series $r_A$
(resp. $r_{A'}$). For any word $w$ let
$\theta(w)=(\mu(w)\gamma, \mu'(w)\gamma')$. Let $E$ be the vector
subspace of ${\mathbb R}^{n+n'}$ spanned by $\{\theta(w)|w\in
\Sigma^*\}$ and let $T$ be the linear mapping from ${\mathbb
  R}^{n+n'}$ to ${\mathbb R}$ defined by $T(u,u')=\lambda u-\lambda'
u'$ for any $u\in {\mathbb R}^n$ and $u'\in {\mathbb R}^{n'}$. The
series $r_A$ and $r_{A'}$ are equal, i.e. $ A$ and $A'$ are equivalent, iff $\forall (u,u')\in E, T(u,u')=0$,
property which can be checked within polynomial time. \qed
\end{proof}
The following algorithm decides the equivalence of two MA:

\bigskip

\noindent \texttt{Input: $A$, $A'$ MA}

\noindent \texttt{$B=\{\varepsilon\}, S=\{x|x\in \Sigma\}$}

\noindent \texttt{while $S\neq \emptyset$ do}

\noindent \texttt{~~~~let $v$ be the smallest element in $S$ and let $S=S\setminus \{v\}$}

\noindent \texttt{~~~~if $\theta(v)$ does not belong to the subspace spanned by $\theta(B)$ then}

\noindent \texttt{~~~~~~~~$B=B\cup \{v\}$ and $S=S\cup\{vx|x\in \Sigma\}$}

\noindent \texttt{~~~~end if}

\noindent \texttt{end while}

\noindent \texttt{while $B \neq \emptyset$ do}

\noindent \texttt{~~~~let $v\in B$ and let $B=B\setminus \{v\}$}

\noindent \texttt{~~~~if $T(\theta(v))\neq 0$ then}

\noindent \texttt{~~~~~~~~output \textbf{no} ; exit}

\noindent \texttt{~~~~end if}

\noindent \texttt{end while}

\noindent \texttt{output \textbf{yes}}.

\bigskip 

The first part of the algorithm computes a basis of $E$; the second part checks whether $T(E)=\{0\}$. 

Note that when $A$ and $A'$ are not equivalent, the previous algorithm
provides a word $u$ such that $r_A(u)\neq r_{A'}(u)$ and whose length
is $\leq |Q|+|Q'|$.

\begin{proposition}\label{Prop:spanMA}
  Let $A_0, A_1, \ldots, A_n$ be MAs over ${\mathbb R}$. It is
  decidable within polynomial time whether there exists $\alpha_1,
  \ldots, \alpha_n\in {\mathbb R}$ such that $r_{A_0}=\sum_{i=1}^n
  \alpha_ir_{A_i}$. 
More precisely, all such tuples of parameters
  $(\alpha_1, \ldots, \alpha_n)$ are solutions of a linear system
  computable within polynomial time.
\end{proposition}

\begin{proof}
Consider the following algorithm.

\medskip

\noindent \texttt{Let $Eq=\{r_{A_0}(\varepsilon)=\sum_{i=1}^nx_ir_{A_i}(\varepsilon)\}$}

\noindent \texttt{\#$Eq$ is a set of independent equations on variables $x_1, \ldots, x_n$.}

\noindent \texttt{While $Eq$ has a solution $(\alpha_1, \ldots, \alpha_n)$ such that $r_{A_0}\neq \sum_{i=1}^n  \alpha_ir_{A_i}$}

\noindent \texttt{~~~~Let $u$ be a word such that $r_{A_0}(u) \neq \sum_{i=1}^n  \alpha_ir_{A_i}(u)$}

\noindent \texttt{~~~~$Eq=Eq\cup\{r_{A_0}(u) = \sum_{i=1}^n  x_ir_{A_i}(u)\}$}

\noindent \texttt{Output  : $Eq$}

\bigskip 

From Proposition~\ref{Prop:eqMA}, if $r_{A_0} \neq \sum_{i=1}^n
\alpha_ir_{A_i}$, a word $u$ such that $r_{A_0}(u) \neq \sum_{i=1}^n
\alpha_ir_{A_i}(u)$ and whose length is $\leq \sum_{i=0}^n |Q_i|$ can
be found within polynomial time (where $|Q_i|$ is the number of states
of $A_i$). The algorithms ends since $Eq$ has at most $n+1$
elements. It is clear that $(\alpha_1, \ldots, \alpha_n)$ is a
solution of $Eq$ iff $r_{A_0} = \sum_{i=1}^n \alpha_ir_{A_i}$.  \qed
\end{proof}

A similar result holds when we ask for positive coefficients.

\begin{proposition}\label{Prop:spanMA+}
  Let $A_0, A_1, \ldots, A_n$ be MAs over ${\mathbb R}$. It is
  decidable within polynomial time whether there exists $\alpha_1,
  \ldots, \alpha_n\in {\mathbb R}^+$ such that $r_{A_0}=\sum_{i=1}^n
  \alpha_ir_{A_i}$. 
\end{proposition}
\begin{proof}
Add the constraints $x_1\geq 0, \ldots, x_n\geq 0$ to the system $Eq$
in the previous algorithm. A polynomial linear programming algorithm
will then find a solution of $Eq$ or decide that $Eq$ has no solution.\qed
\end{proof}

As a consequence of these propositions, it can efficiently be decided whether an MA
$A$ is $K$-reduced
.
\begin{proposition}\label{Prop:Kreduced}
 Let $A=\left\langle \Sigma,Q,\varphi,\iota,\tau\right\rangle $ be a $K$-MA. It is decidable within polynomial time whether $A$ is $K$-reduced; if $A$ is
 not $K$-reduced, a $K$ reduction can be computed within polynomial
 time.
\end{proposition}
\begin{proof}
For any $q\in Q$, check whether there exist coefficients
$\alpha_{q'}\in K$ for $q'\in Q'=Q\setminus\{Q\}$ such that
$r_{A,q}=\sum_{q'\in Q'}\alpha_{q'}r_{A,q'}$. If so, use these
coefficients to compute a $K$-reduction of $A$.\qed
\end{proof}

\section{Rational stochastic languages}\label{rsl}

The objects we study are \emph{rational stochastic languages}, i.e.
stochastic languages which are also rational series. A rational stochastic
language can always be generated by using a multiplicity
automaton. But depending on the set $K$ of numbers used for the
parameters, we obtain different sets ${\cal S}_K^{rat}(\Sigma)$ of
rational stochastic languages. In the following, we
suppose that $K\in \{{\mathbb R}, {\mathbb R^+}, {\mathbb Q}, {\mathbb
  Q^+}\}$. First, we study the relations between all these classes of
  rational stochastic languages and next, we give a
  characterization of ${\cal S}_K^{rat}(\Sigma)$ in terms of stable
  subsemimodules of  ${\cal S}(\Sigma)$.

\subsection{Relations between classes of rational stochastic languages}\label{crsl}

Let us begin by the simplest inclusions. 

\begin{proposition}
 $${\cal S}_{\mathbb Q^+}^{rat}(\Sigma) \subseteq{\cal S}_{\mathbb Q}^{rat}(\Sigma) \subsetneq {\cal
S}_{\mathbb R}^{rat}(\Sigma) \textrm{ and }{\cal S}_{\mathbb Q^+}^{rat}(\Sigma) \subsetneq {\cal S}_{\mathbb
R^+}^{rat}(\Sigma)\subseteq {\cal S}_{\mathbb
R}^{rat}(\Sigma). $$
Moreover, $${\cal S}_{\mathbb
R^+}^{rat}(\Sigma)\setminus {\mathbb Q}\langle\langle
  \Sigma\rangle\rangle \neq \emptyset.$$
\end{proposition}
\begin{proof}
Let $K_1$ be a subsemiring of $K_2$. We have $K_1^{rat}\langle\langle 
  \Sigma\rangle\rangle\subseteq K_2^{rat}\langle\langle 
  \Sigma\rangle\rangle$ and hence, ${\cal S}_{K_1}^{rat}(\Sigma) \subseteq{\cal S}_{K_2}^{rat}(\Sigma)$. 

Now, let $r$ be the rational series defined on $\Sigma=\{a\}$ by
$r(\varepsilon)=\sqrt{2}/2, r(a)=1-\sqrt{2}/2$ and $r(a^n)=0$ for any
$n\geq 2$. Clearly, $r\in {\cal S}_{\mathbb
R^+}^{rat}(\Sigma)\setminus {\mathbb Q}\langle\langle
\Sigma\rangle\rangle$ which implies that ${\cal S}_{\mathbb
Q}^{rat}(\Sigma) \subsetneq {\cal S}_{\mathbb R}^{rat}(\Sigma)$ and
${\cal S}_{\mathbb Q^+}^{rat}(\Sigma) \subsetneq {\cal S}_{\mathbb
R^+}^{rat}(\Sigma)$.\qed
\end{proof}

A rational stochastic language over ${\mathbb R}$ which only takes rational
values is a rational stochastic language over ${\mathbb Q}$.

\begin{proposition}\label{prop:fatou}$${\cal S}_{\mathbb R}^{rat}(\Sigma)\cap {\mathbb Q}\langle\langle
  \Sigma\rangle\rangle={\cal S}_{\mathbb Q}^{rat}(\Sigma).$$
\end{proposition}

\begin{proof}
\end{proof}
Recall that ${\mathbb R}$ is a Fatou
extension of ${\mathbb Q}$: any rational series over
${\mathbb R}$ which only takes rational values is a rational
series over ${\mathbb Q}$ i.e. 
$${\mathbb R}^{rat}\langle\langle
  \Sigma\rangle\rangle \cap {\mathbb Q}\langle\langle
  \Sigma\rangle\rangle = {\mathbb Q}^{rat}\langle\langle
  \Sigma\rangle\rangle.$$ 
As a consequence, 
\begin{align*}
{\cal S}_{\mathbb R}^{rat}(\Sigma)\cap {\mathbb Q}\langle\langle
  \Sigma\rangle\rangle&={\cal S}(\Sigma)\cap{\mathbb R}^{rat}\langle\langle
  \Sigma\rangle\rangle\cap {\mathbb Q}\langle\langle
  \Sigma\rangle\rangle\\
&={\cal S}(\Sigma)\cap{\mathbb Q}^{rat}\langle\langle
  \Sigma\rangle\rangle\\
&={\cal S}_{\mathbb Q}^{rat}(\Sigma).
\end{align*}\qed

It has also been proved that ${\mathbb
R^+}$ is not a Fatou extension of ${\mathbb
Q^+}$: ${\mathbb
Q^+}^{rat}\langle\langle \Sigma\rangle\rangle \subsetneq {\mathbb
R^+}^{rat}\langle\langle \Sigma\rangle\rangle\cap {\mathbb
Q^+}\langle\langle \Sigma\rangle\rangle$. We prove below that this
result can be extended to stochastic languages: there exists a rational stochastic
language over ${\mathbb R^+}$ which takes only rational values and
which is not a rational stochastique language over ${\mathbb Q^+}$.

\begin{proposition}
${\cal S}_{\mathbb Q^+}^{rat}(\Sigma) \subsetneq {\cal S}_{\mathbb R^+}^{rat}(\Sigma)\cap {\mathbb Q^+}\langle\langle
  \Sigma\rangle\rangle$.
\end{proposition}
\begin{proof}
  We use an element in ${\mathbb R^+}^{rat}\langle\langle
  \Sigma\rangle\rangle\cap {\mathbb Q^+}\langle\langle
  \Sigma\rangle\rangle\setminus {\mathbb Q^+}^{rat}\langle\langle
  \Sigma\rangle\rangle$ described in \cite{BerstelReutenauer84} to prove
  the proposition.

Consider the multiplicity automaton $A=\left\langle
    \Sigma,Q,\varphi,\iota,\tau\right\rangle $ where $\Sigma=\{a,b\}$,
    $Q=\{q_0,q_1\}$, $\iota(q_0)=\iota(q_1)=1$,
    $\varphi(q_0,a,q_0)=\alpha^2$, $\varphi(q_0,b,q_0)=\alpha^{-2}$,
    $\varphi(q_1,a,q_1)=\alpha^{-2}$, $\varphi(q_1,b,q_1)=\alpha^2$ where $\alpha=(\sqrt{5}+1)/2$,
    $\varphi(q_i,x,q_j)=0$ for any $x\in \Sigma$ when $i\neq j$ and
    $\tau(q_0)= \tau(q_1)=1$ (see Figure~\ref{fig:nonFatou}). 

Let $r_A$ be the rational series generated by $A$.  Let $w\in
    \Sigma^*$. We have $r_A(w)=\alpha^{2n}+\alpha^{-2n}$ where
    $n=|w|_a-|w|_b$. Check that for any integer $n$,
    $\alpha^{2n}+\alpha^{-2n}\in {\mathbb N}$. Hence, $r_A\in {\mathbb
    R^+}^{rat}\langle\langle \Sigma\rangle\rangle\cap {\mathbb
    Q^+}\langle\langle \Sigma\rangle\rangle$.  It is shown
    in~\cite{BerstelReutenauer84} that $r_A\not \in {\mathbb
    Q^+}^{rat}\langle\langle \Sigma\rangle\rangle$.
    
    Now let $A'=\left\langle
    \Sigma,Q,\varphi',\iota',\tau'\right\rangle $ where for any states
    $q$ and $q'$ and any letter $x$, $\iota'(q)=1/2$,
    $\varphi'(q,x,q')=\varphi(q,x,q')/4$ and $\tau'(q_0)=
    \tau'(q_1)=1/4$.  Check that $\alpha^2+\alpha^{-2}=3$. Then, $A'$
    is a probabilistic automaton. Let $p$ be the stochastic language
    generated by $A$. We have $$p(w)=\frac{1}{2^{2|w|+3}}\left(
    \alpha^{2n}+\alpha^{-2n}\right)\textrm{ where }n=|w|_a-|w|_b$$
and hence $$p\in {\cal
    S}_{\mathbb R^+}^{rat}(\Sigma)\cap {\mathbb Q^+}\langle\langle
  \Sigma\rangle\rangle.$$

    Let $s$ be the series defined by $s(w)=2^{2|w|+3}$. Clearly, $s\in
    {\mathbb Q^+}^{rat}\langle\langle \Sigma\rangle\rangle$ and
    $r_A=s\odot p$ (Hadamard product). Recall that when $K$ is
    commutative, the Hadamard product of two rational series is a
    rational series. Therefore $r_A\not \in {\mathbb
    Q^+}^{rat}\langle\langle \Sigma\rangle\rangle \Rightarrow p\not
    \in {\mathbb Q^+}^{rat}\langle\langle \Sigma\rangle\rangle$ and
    hence, $p\not \in {\cal S}_{\mathbb Q^+}^{rat}(\Sigma)$.\qed
    
    \end{proof}
  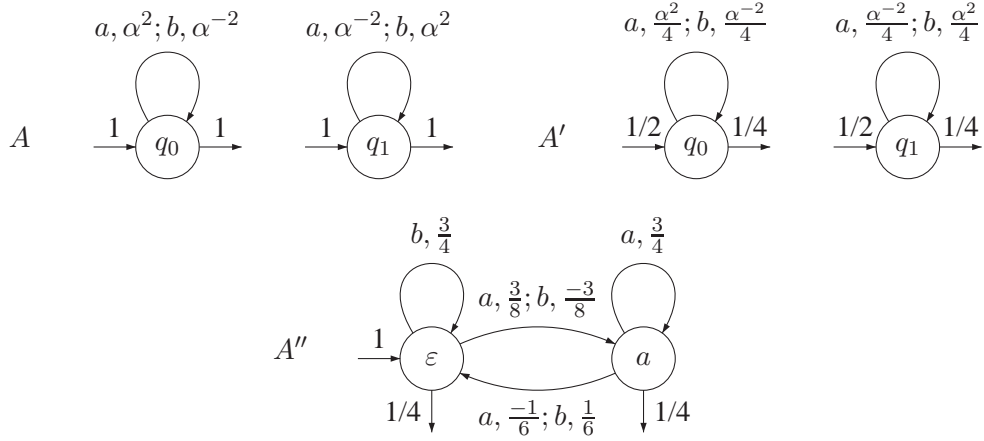
\begin{figure}[h]
    \begin{center}
       \begin{picture}(120,50)
\put(5,35){$A$}
\letstate p0=(10,35)     
\letstate q0=(20,35)      
\letstate t0=(30,35) 
\drawstate(q0){$q_0$}
\drawtrans(p0,q0){1}
\drawtrans(q0,t0){1}
\drawloop[t](q0){$a,\alpha^2;b,\alpha^{-2}$}   
\letstate p1=(30,35)      
\letstate q1=(40,35)      
\letstate t1=(50,35) 
\drawtrans(p1,q1){1}
\drawtrans(q1,t1){1}
\drawstate(q1){$q_1$}
\drawloop[t](q1){$a,\alpha^{-2};b,\alpha^2$}
\put(55,35){$A'$}
\letstate p2=(60,35)     
\letstate q2=(70,35)      
\letstate t2=(80,35) 
\drawstate(q2){$q_0$}
\drawtrans(p2,q2){1/2}
\drawtrans(q2,t2){1/4}
\drawloop[t](q2){$a,\frac{\alpha^2}{4};b,\frac{\alpha^{-2}}{4}$}   
\letstate p3=(80,35)      
\letstate q3=(90,35)       
\letstate t3=(100,35) 
\drawtrans(p3,q3){1/2}
\drawtrans(q3,t3){1/4}
\drawstate(q3){$q_1$}
\drawloop[t](q3){$a,\frac{\alpha^{-2}}{4};b,\frac{\alpha^2}{4}$}

\put(30,15){$A''$}
\letstate p4=(35,15)     
\letstate q4=(45,15)        
\letstate t4=(45,5)      
\letstate q5=(65,15)       
\letstate t5=(65,5) 
\drawstate(q4){$\varepsilon$}  
\drawtrans[r](q4,t4){1/4}  
\drawstate(q5){$a$}  
\drawtrans(q5,t5){1/4} 
\drawtrans(p4,q4){1}
\drawloop[t](q4){$b,\frac{3}{4}$}   
\drawcurvedtrans(q4,q5){$a,\frac{3}{8};b,\frac{-3}{8}$}  
\drawcurvedtrans(q5,q4){$a,\frac{-1}{6};b,\frac{1}{6}$}
\drawloop[t](q5){$a,\frac{3}{4}$} 
\end{picture}
\caption{$A'$ generates a rational stochastic language $p_{A'}$   which takes all its values in ${\mathbb Q}$. However, $p_{A'}$ is not a rational stochastic language over ${\mathbb Q}^+$. $A''$ is a multiplicity automaton over ${\mathbb Q}$ which generates $p_{A'}$.}  
\label{fig:nonFatou}
\end{center}
  \end{figure}

Remark that since $p$ is a rational stochastic language which
takes all its values in ${\mathbb Q}$, $p$ is a rational stochastic
language over ${\mathbb Q}$, from Prop~\ref{prop:fatou}. Let $p_0=p_{A',q_0}$ and $p_1=p_{A',q_1}$ be the stochastic languages generated from the states $q_0$ and $q_1$ of automaton $A'$. It can easily be shown that 
$$\left\{
\begin{array}{ll}
p=\frac{1}{2}p_0+\frac{1}{2}p_1\\
a^{-1}p=\frac{\alpha^2}{3}p_0+\frac{\alpha^{-2}}{3}p_1\\
\end{array}
\right.$$
These relations makes it possible to base on $p$ and $a^{-1}p$ an automata which recognizes $p$.  Check that $$\dot{a}p=\frac{3}{8}p, \dot{b}p=\frac{3}{4}p-\frac{3}{8}a^{-1}p, \dot{a}a^{-1}p=\frac{-1}{6}p-\frac{3}{4}a^{-1}p\textrm{ and }\dot{b}a^{-1}p=\frac{1}{6}p+\frac{3}{4}a^{-1}p.$$
These relations can be used to prove that the automaton $A''$ in Fig.~\ref{fig:nonFatou} generates $p$. 


Now, we prove that there exists a rational stochastic language over ${\mathbb Q}$ which
is not rational over ${\mathbb R^+}$. In particular, it
cannot be generated by a probabilistic automaton.

\begin{proposition}\label{prop:nonFatou}
${\cal S}_{\mathbb Q}^{rat}(\Sigma)\setminus  {\cal S}_{\mathbb R^+}^{rat}(\Sigma)\neq \emptyset$. 
\end{proposition}

\begin{proof}
Let $\Sigma=\{a,b\}$ and for any $w\in \Sigma^*$, let $r$ and $s$ be
the series defined by $r(w)=|w|_a$ and $s(w)=|w|_b$. They are rational
over ${\mathbb Q}$ since they belong to a stable finitely generated
subsemimodule of ${\mathbb Q}\langle\langle\Sigma
\rangle\rangle$. Indeed, $$\dot{a}r=r+1, \dot{b}r=r,
\dot{a}s=s\textrm{ and } \dot{b}s=s+1.$$ Hence, the series $r-s$ and
$(r-s)^2$ where the exponent refers to the Hadamard product are also
rational over ${\mathbb Q}$. For any $n\in {\mathbb N}$, let
$\sigma_n=\sum_{w\in \Sigma^n}(r-s)^2(w)\leq n^2\cdot 2^n$. Check that
$$\sigma_n=n2^n \textrm{ and }\sigma=\sum_{n\geq
0}\frac{\sigma_n}{2^{2n}}=2.$$ 

Now, let $t$ be the series defined by
$$t(w)=\frac{(r-s)^2(w)}{\sigma\cdot 2^{2|w|}}.$$ $t$ is a rational
stochastic languages over ${\mathbb Q}$. Its support is the set
$supp(t)=\{w\in \Sigma^*\ |\ |w|_a\neq |w|_b\}$ which is known to be not
rational. If $t$ were rational over ${\mathbb R}^+$, it support would
be rational. Therefore, $t\in {\cal S}_{\mathbb Q}^{rat}(\Sigma)\setminus  {\cal S}_{\mathbb R^+}^{rat}(\Sigma)$. \qed
\end{proof}

\begin{figure}[htbp]
  \input{recapitulatif1.pstex_t}
  \caption{Inclusion relations between classes of rational stochastic languages.}
\label{recapitulatif1}
  \end{figure}
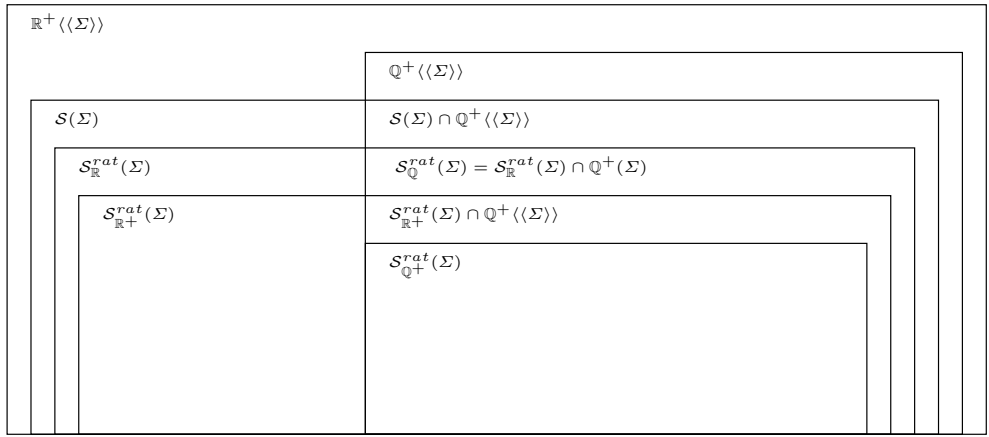

All these results can be summarized on diagram~\ref{recapitulatif1}.

\subsection{Residual languages of rational stochastic languages}\label{rlrsl}

Recall that given a stochastic language $p\in {\cal S}(\Sigma)$ and a
word $u\in res(p)$, i.e.  such that $p(u\Sigma^*)\neq 0$, the
residual language of $p$ wrt $u$ is the stochastic language defined by
$$u^{-1}p(w)=\frac{p(uw)}{p(u\Sigma^*)}\cdot$$ When $p$ takes
its values in ${\mathbb Q^+}$, it is not true in general that
$u^{-1}p$ takes also its values in ${\mathbb Q^+}$.

Consider two series
$(\alpha_n)_{n\in {\mathbb N}}$ and $(\beta_n)_{n\in {\mathbb N}}$
over ${\mathbb Q^+}$ and such that $\sum_n\alpha_n=\sqrt{2}/2$ and
$\sum_n\beta_n=4/5-\sqrt{2}/2$. Now, consider the series $r\in
{\mathbb Q}^+\langle\langle\{a,b\}\rangle\rangle$ defined by
$r(\varepsilon)=1/5$, $r(a^n)=\alpha_{n-1}$, $r(b^n)=\beta_{n-1}$ for
$n\geq 1$ and $r(w)=0$ otherwise. It is easy to check that $r$ is a
stochastic language which takes its values over ${\mathbb Q^+}$ and
that $a^{-1}r(\varepsilon)=\sqrt{2}\alpha_0$. Therefore, $a^{-1}r\not
\in {\mathbb Q}\langle\langle\Sigma\rangle\rangle$.

We prove below that when $p$ is a rational stochastic language over
$K$, all its residual languages are also rational over $K$.
Moreover, the set $Res(p)=\{u^{-1}p|u\in res(p)\}$ generates the
same subsemimodule of $K\langle\langle \Sigma\rangle\rangle$ as the
set $\{\dot{u}p | u\in \Sigma^*\}$.

We need before two linear algebra technical lemmas to prove this result.
\begin{lemma}\label{alglin1}
Let $f: {\mathbb Q}^n\rightarrow {\mathbb Q}^n$ be a linear mapping
and let $\mathbf{t}\in {\mathbb Q}^{n}$ such that $\sum_{k\geq
0}f^k\mathbf{t}$ converges to $\mathbf{u}$. Then $\mathbf{u}\in
{\mathbb Q}^{n}$.
\end{lemma}
\begin{proof}
Let $F$ be the vector subspace of ${\mathbb Q}^n$ generated by
$\{f^k\mathbf{t} | k\in {\mathbb N}\}$. There exists an integer $d$
such that $f^0\mathbf{t}=\mathbf{t}, \ldots, f^{d-1}\mathbf{t}$ is a
basis of $F$. As the sum $\sum_{k\geq 0}f^k\mathbf{t}$ converges,
$f^k\mathbf{t}$ converges to 0 when $k$ tends to infinity. Therefore,
for any $\mathbf{v}\in F$, $f^k\mathbf{v}$ also converges to 0 when
$k$ tends to infinity. Let $\mathbf{v}\in F$ such that
$f\mathbf{v}=\mathbf{v}$. We have also $f^k\mathbf{v}=\mathbf{v}$ for
any integer $k$ and hence, $\mathbf{v}=0$. Let $g: F\rightarrow F$
defined by $g(\mathbf{v})=\mathbf{v}-f\mathbf{v}$. The linear mapping
$g$ is one-to-one and for any $\mathbf{v}\in F$ and any integer $k$,
$$\mathbf{v}+ f\mathbf{v} + \ldots +
f^k\mathbf{v}=g^{-1}(1-f^{k+1})(\mathbf{v}).$$ Therefore,
$$\mathbf{u}=g^{-1}\mathbf{t}\textrm{ and }\mathbf{u}\in {\mathbb
Q}^{n}.$$
\end{proof}

We use Lemma~\ref{alglin1} to show that if $\{r_1, \ldots, r_n\}$ generates
a stable subsemimodule of ${\mathbb Q}\langle\langle
\Sigma\rangle\rangle$ and if each sum $\sum_{w\in \Sigma}r_i(w)$
converges to $\sigma_i$ then each $\sigma_i\in {\mathbb Q}$.

\begin{lemma}\label{sumrat}
Let $M$ be a stable subsemimodule of ${\mathbb Q}\langle\langle
\Sigma\rangle\rangle$ generated by $\{r_1, \ldots, r_n\}$ and let
$\sigma_i^k=\sum_{w\in\Sigma^k}r_i(w)$ for any $1\leq i \leq n$ and
any integer $k$. Suppose that for any $1\leq i \leq n$, the sums
$\sum_{k\geq 0}\sigma_i^k$ converges to $\sigma_i$. Then $\sigma_i\in
{\mathbb Q}$ for any $1\leq i \leq n$.
\end{lemma}
\begin{proof}
Let $\mathbf{t}=(r_1(\varepsilon), \ldots,(r_n(\varepsilon)))^t$. As $M$ is
stable, there exist $\alpha_{i,j}^x\in {\mathbb Q}$ for any $1\leq
i,j\leq n$ and any $x\in \Sigma$ such that $\dot{x}r_i=\sum_{j=1}^n
\alpha_{i,j}^x\cdot r_j$. Let $B\in {\mathbb Q}^{n\times n}$ defined
by $B[i,j]=\sum_{x\in \Sigma}\alpha_{i,j}^x$. Let us prove by
induction on $k$ that for any integer $k$, we have
$(\sigma_1^k,\ldots, \sigma_n^k)^t=B^k\mathbf{t}$. The property is true for
$k=0$ as for any integer $i$, $\sigma_i^0=r_i(\varepsilon)$. Now,
\begin{align*}
  \sigma_i^{k+1}&=\sum_{w\in \Sigma^k,x\in \Sigma}r_i(xw)\\
  &=\sum_{w\in \Sigma^k,x\in \Sigma}\dot{x}r_i(w)\\
  &=\sum_{w\in \Sigma^k, x\in \Sigma, j\in \{1,\ldots, n\}}\alpha_{i,j}^x\cdot r_j(w)\\
  &=\sum_{j\in \{1,\ldots, n\}}\left(\sum_{x\in  \Sigma}\alpha_{i,j}^x\right) \cdot \sum_{w\in \Sigma^k}r_j(w)\\
  &=\sum_{j\in \{1,\ldots, n\}}B[i,j]\sigma_j^k\\
  &=\sum_{j\in \{1,\ldots, n\}}B[i,j](B^k\mathbf{t})[j]\textrm{ by induction hypothesis}\\
  &=(B^{k+1}\mathbf{t})[i].
\end{align*}

Therefore, $B^k\mathbf{t}$ converges to $(\sigma_1,\ldots, \sigma_n)^t$. From
Lemma~\ref{alglin1}, $\sigma_i\in {\mathbb Q}$ for any $1\leq i \leq
n$.\qed
\end{proof}

\begin{lemma}\label{resconv}
Let $p\in {\cal S}_K^{rat}(\Sigma)$. For any word $u\in res(p)$,
$\sum_{w\in \Sigma^*}p(uw)\in K$. Moreover, the set $Res(p)$
generates the same subsemimodule of $K\langle\langle
\Sigma\rangle\rangle$ as the set $\{\dot{u}p | u\in \Sigma^*\}$. 
\end{lemma}

\begin{proof}
Let $p \in {\cal S}_K^{rat}(\Sigma)$. For any word u, $\sum_{w\in
\Sigma^*}p(uw)\in {\mathbb R}^+$ since $p$ is a stochastic
language. Suppose now that $K={\mathbb Q}$ or $K={\mathbb Q}^+$. The
set $\{\dot{u}p | u\in \Sigma^*\}$ generates a finite vector subspace
${\cal P}$ of ${\mathbb Q}\langle\langle \Sigma\rangle\rangle$. Let
$\{\dot{u}_1p, \ldots, \dot{u}_np\}$ be a finite subset of $\{\dot{u}p
| u\in \Sigma^*\}$ which generates ${\cal P}$. Let
$\sigma_i=\sum_{w\in \Sigma^*}\dot{u}_ip(w)$ for any $i=1, \ldots,
n$. From Lemma~\ref{sumrat}, each $\sigma_i\in {\mathbb Q}$. Now, for
any $u\in \Sigma^*$, there exists $\alpha_1, \ldots ,\alpha_n \in
{\mathbb Q}$ such that $\dot{u}p=\sum_{i=1}^n\alpha_i\dot{u}_ip$. Therefore, $\sum_{w\in \Sigma^*}p(uw)=\sum_{i=1}^n\alpha_i\sigma_i\in
{\mathbb Q}^+$.

So, for any $K$ and any $u\in res(p)$, there exists an inversible element $\alpha_u$ of $K$ such that
$\dot{u}p=\alpha_u\cdot u^{-1}p$. In consequence, the set $Res(p)$ generates the same subsemimodule of $K\langle\langle
\Sigma\rangle\rangle$ as the set $\{\dot{u}p | u\in \Sigma^*\}$.  \qed
\end{proof}

For any stochastic language $p$ over $K$, let us denote by $[Res(p)]$
the subsemimodule of $K\langle\langle \Sigma\rangle\rangle$ generated
by $Res(p)$ and let us call it the \emph{residual subsemimodule} of $p$. Note
that $[Res(p)]$ is stable.

\begin{proposition}\label{resrat}
Let $p\in {\cal S}_K^{rat}(\Sigma)$. For any word $u\in res(p)$, $u^{-1}p\in {\cal S}_K^{rat}(\Sigma)$. 
\end{proposition}
\begin{proof}
From Lemma~\ref{resconv}, the residual stochastic languages \ $u^{-1}p$ belong to the same stable subsemimodules of $K\langle\langle
\Sigma\rangle\rangle$ as $p$. Therefore, they are rational over $K$. \qed
\end{proof}

\subsection{Characterization of ${\cal S}_K^{rat}(\Sigma)$ in terms of
stable subsemimodules}\label{charrsl}

We show in this section that a series $p$ over $K$ is a
rational stochastic language if and only if there exists a finite
subset $S$ in  ${\cal S}(\Sigma)$ which generates a stable subsemimodule
of $K\langle\langle \Sigma\rangle\rangle$ and such that $p\in conv_K(S)$.

The « if part » is easy to prove.

\begin{proposition}\label{ifpart}
Let $p\in K\langle\langle
\Sigma\rangle\rangle$. Suppose that there exists a finite
subset $S$ in  ${\cal S}(\Sigma)$ which generates a stable subsemimodule
of $K\langle\langle \Sigma\rangle\rangle$ and such that $p\in conv_K(S)$. Then $p\in {\cal S}_K^{rat}(\Sigma)$.
\end{proposition}

\begin{proof}
  Let $\{p_1, \ldots, p_n\}$ be a finite subset of ${\cal S}(\Sigma)$
  which generates a stable subsemimodule of $K\langle\langle
  \Sigma\rangle\rangle$ and let $p=\sum_{i=1}^n\alpha_ip_i$ where
  $\alpha_i\geq 0$ for $i=1, \ldots, n$ and
  $\sum_{i=1}^n\alpha_i=1$. From Theorem~\ref{caract}, $p$ is a
  rational series over $K$ and $p$ is a stochastic language since
  $p(w)=\sum_{i=1}^n\alpha_ip_i(w)\geq 0$ for any word $w$ and
  $p(\Sigma^*)=\sum_{i=1}^n\alpha_ip_i(\Sigma^*)=1$.\qed
\end{proof}

The converse proposition is easy to prove when $K={\mathbb Q}$ or
$K={\mathbb R}$. It is slightly more complicated when $K$ is not a field.

\begin{proposition}\label{iffpart}
Let $p\in {\cal S}_K^{rat}(\Sigma)$. Then there exists a finite
subset $S$ in  ${\cal S}(\Sigma)$ which generates a stable subsemimodule
of $K\langle\langle \Sigma\rangle\rangle$ and such that $p\in
conv_K(S)$. 
\end{proposition}

\begin{proof}
Let $p\in {\cal S}_K^{rat}(\Sigma)$. 

When $K={\mathbb Q}$ or $K={\mathbb R}$, $K$ is a commutative field,
$K\langle\langle \Sigma\rangle\rangle$ is a vector space and
subsemimodules of $K\langle\langle \Sigma\rangle\rangle$ are vector
subspaces of $K\langle\langle \Sigma\rangle\rangle$. From
Lemma~\ref{resconv}, the subspaces generated by $\{\dot{u}p|u \in
\Sigma^*\}$ and $\{u^{-1}p|u\in \Sigma^*\}$ coincide. From Theorem~\ref{caract}, $\{u^{-1}p|u\in
\Sigma^*\}$ generates a stable finite vector subspace ${\cal P}$ of $K\langle\langle
\Sigma\rangle\rangle$. Let $S$ be a finite subset of $\{u^{-1}p|u\in
res(p)\}$ which contains $p$ and generates ${\cal P}$. Clearly, $S\subseteq
{\cal S}(\Sigma)$ and $p\in conv_K(S)$. 

Let $K={\mathbb Q}^+$ or $K={\mathbb R}^+$.
From Theorem~\ref{caract}, let $R=\{r_1, \ldots, r_n\}$ be a finite
subset of $K\langle\langle \Sigma\rangle\rangle$ which generates a
stable subsemimodule $M$ containing $p$.  We may suppose that
$0\not\in R$ as $R$ and $R\setminus \{0\}$ generate the same
subsemimodule. Let $S=\{r\in R | \sum_{w\in \Sigma^*}r(w)<\infty\}$.
First, let us show that $S$ also generates a stable subsemimodule
containing $p$.  Let $T=R\setminus S$. Let $s\in S$ and let $u\in
\Sigma^*$.  As $M$ is stable, we can write $\dot{u}s=\sum_{r\in
R}\alpha_r^ur$, where the coefficients $\alpha_r^u$ belong to $K$.  As
$s\in S$, $\sum_{w\in \Sigma^*}\dot{u}s(w) < \infty$.  Therefore,
$r\in T \Rightarrow \alpha_r^u=0$ and $S$ generates a stable
subsemimodule.  In a similar way, we can write $p=\sum_{r\in
R}\beta_rr$ and as $p$ is a stochastic language, $r\in T \Rightarrow
\alpha_r=0$ and $p$ belongs to the semimodule generated by $S$.

Now, let $S'=\{\left(\sum_ {w\in \Sigma^*}s(w)\right)^{-1}\cdot s | s
\in S\}$. Clearly, each element of $S'$ is a stochastic language and
an element of $K\langle\langle \Sigma\rangle\rangle$ ( by using
Lemma~\ref{sumrat} when $K={\mathbb Q}^+$). $S'$ generates
the same stable semimodule as $S$.
We can write $p=\sum_{s\in S'}\beta_ss$, where the coefficients $\beta_s$ belong to $K$.  As $p$ and each element of $S'$ is a stochastic language, we have $\sum_{s\in S'}\beta_s=1$ and hence, $p\in conv_K(S')$.\qed
\end{proof}

Putting together the previous propositions, we obtain the following theorem:

\begin{theorem}\label{caracS+}
Let $K\in \{{{\mathbb R}, {\mathbb Q}, \mathbb R}^+, {\mathbb Q}^+\}$.
A series $p$ over $K$ is a rational stochastic language if
and only if there exists a finite subset $S$ in ${\cal S}(\Sigma)$
which generates a stable subsemimodule of $K\langle\langle
\Sigma\rangle\rangle$ and such that $p\in conv_K(S)$.
\end{theorem}

\begin{proof}
Apply Propositions~\ref{ifpart} and \ref{iffpart}.\qed
\end{proof}

\subsection{Subclasses of rational languages defined in terms of properties of their set of residual languages}\label{fingenfin}
 
Let $p$ be a rational stochastic language over $K$. The set $Res(p)$
composed of the stochastic residual languages of $p$ is
\emph{included} in a stable finitely generated subsemimodule of
$K\langle\langle \Sigma\rangle\rangle$ but it may happen that the
residual subsemimodule 
$[Res(p)]$ of $p$  \emph{is not} finitely
generated. See Example~\ref{ex:nonfingen} for instance. In the
opposite, a stochastic language whose residual subsemimodule is
finitely generated is rational. Therefore, two
subclasses of ${\cal S}_K^{rat}(\Sigma)$ can be naturally defined:
\begin{itemize}
\item the set ${\cal S}_K^{fingen}(\Sigma)$ composed of rational
stochastic languages over $K$ whose residual subsemimodule is finitely generated;
\item the set ${\cal S}_K^{fin}(\Sigma)$ composed of rational
stochastic languages over $K$ such that $Res(p)$ is finite.
\end{itemize}

\subsubsection{Stochastic languages with finitely many residual languages.}

Every stochastic languages with finitely many residual languages can
be described by using positive parameters only. In consequence, we obtain a
Fatou-like property: every stochastic language with finitely many
residual languages and which takes its values in ${\mathbb Q}$ is
rational over ${\mathbb Q}^+$. Of course, for any $K$, there
exist rational stochastic languages over $K$ whose residual subsemimodule is
finitely generated and which
have not finitely many residual languages.

\begin{proposition}\label{stofin}
\begin{enumerate}
\item ${\cal S}_{\mathbb R}^{fin}(\Sigma)={\cal S}_{\mathbb R^+}^{fin}(\Sigma)$
\item ${\cal S}_{\mathbb Q}^{fin}(\Sigma)={\cal S}_{\mathbb Q^+}^{fin}(\Sigma)={\cal S}_{\mathbb R}^{fin}(\Sigma)\cap {\mathbb Q}^+\langle\langle \Sigma\rangle\rangle.$
\item For any $K\in \{{\mathbb R}, {\mathbb Q}, {\mathbb R}^+, {\mathbb Q}^+\}$, ${\cal S}_K^{fin}(\Sigma)\subsetneq {\cal S}_K^{fingen}(\Sigma)$.
\end{enumerate}
\end{proposition}
\begin{proof}
\begin{enumerate}
\item It is sufficient to show that ${\cal S}_{\mathbb
    R}^{fin}(\Sigma)\subseteq {\cal S}_{\mathbb R^+}^{fin}(\Sigma)$ in
  order to prove the first equality. Let $p\in {\cal S}_{\mathbb
    R}^{fin}(\Sigma)$ and let $Res(p)=\{u_1^{-1}p, \ldots,
  u_n^{-1}p\}$ be the set of residual languages of $p$. For any $u\in
  \Sigma^*$ and any $i\in \{1, \ldots, n\}$, there exists $j\in \{1,
  \ldots, n\}$ such that
  $\dot{u}u_i^{-1}p=u_i^{-1}p(u\Sigma^*)u_j^{-1}p$. Since
  $u_i^{-1}p(u\Sigma^*)\geq 0$, $Res(p)$ generates a stable
  subsemimodule of ${\mathbb R^+}\langle\langle \Sigma\rangle\rangle$.
  Since $p\in Res(p)$, $p\in {\cal S}_{\mathbb R^+}^{fin}(\Sigma)$ from
  Theorem~\ref{caracS+}.
\item The proof of the first equality goes in a similar way, with the
complementary argument that $u_i^{-1}p(u\Sigma^*)\in {\mathbb
Q}$ from Lemma~\ref{resconv}.

Now, let $p\in {\cal S}_{\mathbb R}^{fin}(\Sigma)\cap {\mathbb Q}^+\langle\langle \Sigma\rangle\rangle.$ From Prop.~\ref{prop:fatou}, $p\in {\cal S}_{\mathbb Q}^{rat}(\Sigma)$. Therefore, $p\in {\cal S}_{\mathbb Q}^{fin}(\Sigma)$.

\item Consider the probabilistic automaton defined on
Fig.~\ref{fig:fingennonfin}. It defines a stochastic language $p$
over ${\mathbb Q}^+$. 
Let us show that $p\in{\cal S}_{{\mathbb
Q}^+}^{fingen}(\Sigma)\setminus {\cal S}_{{\mathbb Q}^+}^{fin}(\Sigma)$ .

\begin{figure}[h]
    \begin{center}
       \begin{picture}(60,25)
\put(5,10){$A$}
\letstate i0=(10,10)     
\letstate q0=(20,10)       
\letstate q1=(40,10)      
\letstate t0=(20,20)  
\drawtrans(i0,q0){1}
\drawstate(q0){$q_0$}
\drawstate(q1){$q_1$}
\drawcurvedtrans(q0,q1){$a,1/2$}
\drawcurvedtrans(q1,q0){$a,1/2$}
\drawloop[t](q1){$a,1/2$} 
\drawtrans(q0,t0){1/2}
\end{picture}
\caption{The automaton $A$ generates a stochastic language over
${\mathbb Q}^+$ whose residual subsemimodule is finitely generated but which has infinitely many residual languages.}\label{fig:fingennonfin}
\end{center}
  \end{figure}
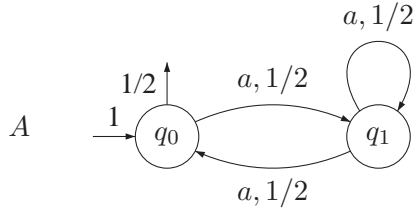

First, let us show by induction on $n$ that for any integer $n$, there exist $\alpha_n, \beta_n\in {\mathbb Q}^+$ such that $\dot{\overline{a^n}}p=\alpha_np+\beta_n\dot{a}p$. This is true when $n=0$: take $\alpha_0=1$ and $\beta_0=0$. Suppose that the relation holds for the integer $n$. For any word $u$, we have:
\begin{align*}
\dot{\overline{a^{n+1}}}p(u)&=\dot{\overline{a^n}}p(au)&\\
&=\alpha_np(au)+\beta_n\dot{a}p(au)\textrm{ by induction hypothesis}\\
&=\frac{\alpha_n}{2}\dot{a}p(u)+\beta_n\left(\frac{1}{2}p(u)+\frac{1}{2}\dot{a}p(u)\right)\textrm{ by remarking that }p=p_{q_0}\\
&\hspace*{6.7cm}\textrm{ and }\dot{a}p=p_{q_1}.\\
\end{align*}
So we can take $\alpha_{n+1}=\beta_n/2$ and
$\beta_{n+1}=(\alpha_n+\beta_n)/2$ which belong to ${\mathbb Q}^+$
from induction hypothesis. Therefore the module $[Res(p)]$ is finitely
generated from Lemma~\ref{resconv}: $p\in
{\cal S}_{{\mathbb Q}^+}^{fingen}(\Sigma)$ and therefore, $p\in
{\cal S}_K^{fingen}(\Sigma)$ for any $K\in \{{\mathbb R}, {\mathbb Q},
{\mathbb R}^+, {\mathbb Q}^+\}$.

Let $\gamma_n=(a^n)^{-1}p(\varepsilon)$. We have $$\gamma_n=\frac{\alpha_np(\varepsilon)+\beta_n\dot{a}p(\varepsilon)}{\alpha_n+\beta_n}=\frac{\alpha_n}{2(\alpha_n+\beta_n)}.$$
Check that $\gamma_n$ satisfies the following induction relation:
$$\gamma_{n+1}=\frac{1-2\gamma_n}{4(1-\gamma_n)}.$$ The sequence
$(\gamma_n)$ converges to the irrational number $(3-\sqrt{5})/4$ and
therefore, $\gamma_n=(a^n)^{-1}p(\varepsilon)$ takes an infinite
number of values, which implies that $p$ has infinitely many residual
languages. \qed
\end{enumerate}
\end{proof}

\subsubsection{Stochastic languages whose residual subsemimodule is finitely generated .}

When $K$ is a field, every rational stochastic language is finitely
generated. This property is no longer true when $K\in \{{\mathbb R}^+,
{\mathbb Q}^+\}$. In consequence, some stochastic languages whose
residual subsemimodule is finitely generated cannot be generated by using only
positive parameters.

We prove also a Fatou-like property: every stochastic
language over ${\mathbb R}^+$ whose residual subsemimodule is finitely
generated and which takes its values in ${\mathbb Q}$ is rational over
${\mathbb Q}^+$. But we first need the following technical lemmas.

\begin{lemma}
  Let $k,n\in {\mathbb N}$ and let $\alpha_i, \beta_i^j\in {\mathbb
    Q}$for $1\leq i \leq n$ and $1\leq j \leq k$. Consider the
  variables $x_1, \ldots, x_k$ and the system $(S)$ composed of the
  $n$ following inequations
  $$\alpha_i+\sum_{j=1}^kx_j\beta^j_i\geq 0$$
  for $i=1, \ldots, n$. If
  $(S)$ has a solution, then it has also a solution which satisfies
  $$\alpha_i+\sum_{j=1}^kx_j\beta^j_i\in {\mathbb Q}^+$$
  for $i=1,
  \ldots, n$.
\end{lemma}
\begin{proof}By induction on $n$. \begin{itemize}
\item Let $n=1$. Let $\mu_1, \ldots, \mu_k$ be such that
$\alpha_1+\sum_{j=1}^k\mu_j\beta^j_1\geq 0$. If
$\alpha_1+\sum_{j=1}^k\mu_j\beta^j_1= 0$, we are done. If
$\alpha_1+\sum_{j=1}^k\mu_j\beta^j_1> 0$, there exists $\mu'_1,
\ldots, \mu'_k\in {\mathbb Q}$ such that
$\alpha_1+\sum_{j=1}^k\mu'_j\beta^j_1> 0$ since ${\mathbb Q}$ is dense
in ${\mathbb R}$ and since $\alpha_1+\sum_{j=1}^k\mu_j\beta^j_1$ is a
continuous expression of the $\mu_i$.
\item Let $n>1$ and let $\mu_1, \ldots, \mu_k$ be such that
$\alpha_i+\sum_{j=1}^k\mu_j\beta^j_i\geq 0$ for any $1\leq i \leq
n$. If $\alpha_i+\sum_{j=1}^k\mu_j\beta^j_i> 0$ for any integer $i$,
then there exists $\mu'_1, \ldots, \mu'_k\in {\mathbb Q}$ such that
$\alpha_i+\sum_{j=1}^k\mu'_j\beta^j_i> 0$ for any $i$, by using the
same argument as previously. Otherwise, there exists at least an
integer $i$ such that $\alpha_i+\sum_{j=1}^k\mu_j\beta^j_i= 0$.  \begin{itemize}
\item If
each $\beta^j_i=0$, then $\alpha_i$ is also null and this equation can
be ruled out from the system without modifying its solutions. In this
case, the induction hypothesis can be directly applied. 
\item If there exists $j$ such that $\beta_i^j\neq 0$, then $\mu_j$
can be expressed as a function of the other $\mu_i$:
$\mu_j=-(\alpha_i+\sum_{l\neq j}\mu_l\beta^l_i)/\beta_i^j$, $x_j$ can
be replaced with $-(\alpha_i+\sum_{l\neq j}x_l\beta^l_i)/\beta_i^j$ in
all the other inequations and the induction hypothesis can be applied.
\end{itemize}
\end{itemize}\qed
\end{proof}

\begin{lemma}\label{R+doncQ+}
Let $r_0, r_1, \ldots, r_n\in {\mathbb Q}\langle\langle \Sigma \rangle\rangle$ and let $\alpha_1, \ldots, \alpha_n\in {\mathbb Q}$, $\beta_1, \ldots, \beta_n\in {\mathbb R}^+$ be such that $$r_0=\sum_{i=1}^n\alpha_ir_i=\sum_{i=1}^n\beta_ir_i.$$
Then, there exists $\gamma_1, \ldots, \gamma_n\in {\mathbb Q}^+$ 
such that $$r_0=\sum_{i=1}^n\gamma_ir_i.$$
\end{lemma}
\begin{proof}
The set of parameters $\{(\lambda_1, \ldots, \lambda_n)\in {\mathbb
R}^n | \sum_{i=1}^n\lambda_ir_i=0\}$ is a vector subspace of ${\mathbb
R}^n$. Since the series $r_1, \ldots, r_n$ take their values in
${\mathbb Q}$, there exist $k$ vectors $(t^1_1, \ldots, t^1_n), \ldots,(t^k_1,
\ldots, t^k_n) \in {\mathbb Q}^n$, with $k\leq n$,  such that for any $(\lambda_1,
\ldots, \lambda_n)\in {\mathbb R}^n$, $$\sum_{i=1}^n\lambda_ir_i=0
\textrm{ iff }\exists \mu_1, \ldots, \mu_k\in {\mathbb R} \textrm{
s.t. }\lambda_i=\sum_{j=1}^k\mu_jt^j_i\textrm{ for any }i=1, \ldots, n.$$
Hence,  for any $(\lambda_1,
\ldots, \lambda_n)\in {\mathbb R}^n$, $$r_0=\sum_{i=1}^n\lambda_ir_i
\textrm{ iff }\exists \mu_1, \ldots, \mu_k\in {\mathbb R} \textrm{
s.t. }\lambda_i=\alpha_i+\sum_{j=1}^k\mu_jt^j_i\textrm{ for any }i=1, \ldots, n.$$
In particular, there exist $\mu_1, \ldots, \mu_k$ such that $\beta_i=\alpha_i+\sum_{j=1}^k\mu_jt^j_i\geq 0\textrm{ for any }i=1, \ldots, n.$ 

Consider the system composed of the $n$ inequations
$\alpha_i+\sum_{j=1}^kx_jt^j_i\geq 0$ for $i=1, \ldots, n.$ It has a
solution and from the previous Lemma, it has also a solution $(\mu_1,
\ldots, \mu_k)$ which satisfies $\alpha_i+\sum_{j=1}^k\mu_jt^j_i\in
{\mathbb Q}^+$ for $i=1, \ldots, n.$ \qed
\end{proof}

\begin{proposition}
\begin{enumerate}
\item When  $K\in \{{\mathbb R}, {\mathbb Q}\}$, ${\cal S}_K^{fingen}(\Sigma)= {\cal S}_K^{rat}(\Sigma)$.
\item When $K\in \{{\mathbb R}^+, {\mathbb Q}^+\}$, ${\cal S}_K^{fingen}(\Sigma)\subsetneq {\cal S}_K^{rat}(\Sigma)$.
\item ${\cal S}_{\mathbb Q^+}^{fingen}(\Sigma)={\cal S}_{{\mathbb R}^+}^{fingen}(\Sigma)\cap {\mathbb Q}^+\langle\langle \Sigma\rangle\rangle.$
\end{enumerate}
\end{proposition}

\begin{proof}
\begin{enumerate}
\item When $K\in \{{\mathbb R}, {\mathbb Q}\}$, $K$ is a commutative
field. As a consequence, any vector subspace of a finitely generated
vector subspace of $K\langle\langle \Sigma\rangle\rangle$ is finitely
generated itself. Therefore, for any $p\in {\cal S}_K^{rat}(\Sigma)$,
the residual subsemimodule of $p$ is
finitely generated. 
\item Example~\ref{ex:nonfingen} describes a rational stochastic
      language whose residual subsemimodule is not finitely generated. 
\item Let $p\in {\cal S}_{{\mathbb R}^+}^{fingen}(\Sigma)\cap {\mathbb
Q}^+\langle\langle \Sigma\rangle\rangle.$ Let $S=\{r_1, \ldots,
r_n\}\subseteq Res(p)$ be a finite subset which generates the same
subsemimodule as $Res(p)$ in ${\mathbb R}^+\langle\langle
\Sigma\rangle\rangle.$ From Prop.~\ref{prop:fatou}, $p\in {\cal
S}_{\mathbb Q}^{rat}(\Sigma)$ and from Prop.~\ref{resrat}, each
$r_i\in {\cal S}_{\mathbb Q}^{rat}(\Sigma)$. $S$ also generates the
same subsemimodule as $Res(p)$ in ${\mathbb Q}\langle\langle
\Sigma\rangle\rangle.$ From Lemma~\ref{R+doncQ+}, for any word $u$ and
any index $i$, there exists $\gamma^{i,u}_1, \ldots, \gamma^{i,u}_n\in
{\mathbb Q}^+$ such that $\dot{u}r_i=\sum_{j=1}^n\gamma^{i,u}_jr_j.$
Therefore, $S$ generates a stable subsemimodule of ${\mathbb
Q}^+\langle\langle \Sigma\rangle\rangle.$ Also from
Lemma~\ref{R+doncQ+}, there exists $\gamma_1, \ldots, \gamma_n\in
{\mathbb Q}^+$ such that $p=\sum_{i=1}^n\gamma_ir_i.$ Therefore, $p\in
conv_{{\mathbb Q}^+}(S)$ and $p\in {\cal S}_{\mathbb
Q^+}^{fingen}(\Sigma)$.
\end{enumerate}\qed
\end{proof}

Remark that ${\cal S}_{\mathbb Q^+}^{fingen}(\Sigma)\subsetneq {\cal
S}_{\mathbb R}^{fingen}(\Sigma)\cap {\mathbb Q}^+\langle\langle
\Sigma\rangle\rangle$ since ${\cal S}_{\mathbb
Q^+}^{fingen}(\Sigma)\subsetneq {\cal S}_{\mathbb
Q}^{rat}(\Sigma)={\cal S}_{\mathbb R}^{rat}(\Sigma)\cap {\mathbb
Q}^+\langle\langle \Sigma\rangle\rangle={\cal S}_{\mathbb
R}^{fingen}(\Sigma)\cap {\mathbb Q}^+\langle\langle
\Sigma\rangle\rangle$

Finaly, we show that when $K$ is positive, finitely generated stochastic
languages over $K$ have a unique normal representation in terms of
stable subbsemimodules generated by residual languages which is
minimal with respect to inclusion.

\begin{proposition}\label{primeResidual}
Let $K={\mathbb Q}^+$ or $K={\mathbb R}^+$ and let $p\in {\cal
S}_K^{fingen}(\Sigma)$. Then, there exists a unique finite subset
$R\subseteq Res(p)$ which generates a stable subsemimodule of
$K\langle\langle \Sigma\rangle\rangle$, such that $p\in conv_K(R)$ and which is minimal for inclusion.
\end{proposition}

\begin{proof}
Let $K={\mathbb Q}^+$ or $K={\mathbb R}^+$ and let $p\in {\cal
S}_K^{fingen}(\Sigma)$. Let  $R=\{r_1, \ldots, r_n\}$ and $S=\{s_1,
\ldots, s_m\}$ be two minimal subsets of $Res(p)$ generating $[Res(p)]$. Let
$r_{i_0}\in R$. We are to prove that $r_0\in S$. 

There exist $\alpha_{i_0}^1, \ldots, \alpha_{i_0}^n \in K$ such that
$r_{i_0}=\sum_{i=1}^m \alpha_{i_0}^is_i$.

There exist $\beta_i^j\in K$ for any $1\leq i, j \leq n$
such that $s_i=\sum_{j=1}^n \beta_i^jr_j$ for any $1\leq i\leq m$.

Therefore, $$r_{i_0}=\sum_{i=1}^m \alpha_{i_0}^i\sum_{j=1}^n \beta_i^jr_j=\sum_{j=1}^n \left(\sum_{i=1}^m\alpha_{i_0}^i\beta_i^j\right) r_j.$$

 If $\sum_{i=1}^m\alpha_{i_0}^i\beta_i^{i_0}<1$, then we could express
 $r_{i_0}$ as a convex combination of the other $r_i$ and $R$ would
 not be minimal for inclusion. Therefore,
 $\sum_{i=1}^m\alpha_{i_0}^i\beta_i^{i_0}=1$.

Since $\sum_{i=1}^m \alpha_{i_0}^i=1$ and each $\beta_i^j\in [0,1]$,
for any index $i$ such that $\alpha_{i_0}^i\neq 0$, we must have
$\beta_i^{i_0}=1$. Therefore, for any index $i$ such that
$\alpha_{i_0}^i\neq 0$, we must have $s_i=r_{i_0}$. As such an index
must exist, $r_{i_0}\in S$.

Since no condition has been put on $r_{i_0}$, then $R\subseteq S$ and finally, $R=S$. \qed
\end{proof}

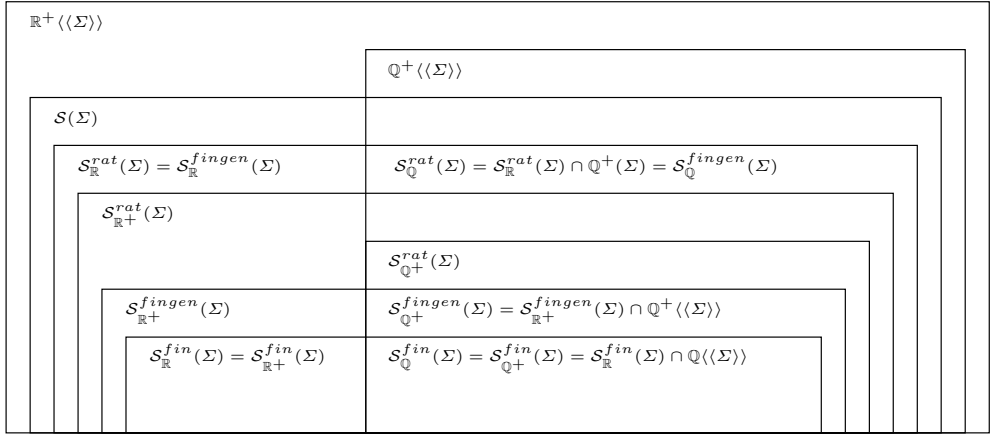
\begin{figure}[htbp]\label{recapitulatif2}
  \input{recapitulatif2.pstex_t}
  \caption{Inclusion relations between classes of classes of rational
  stochastic languages, including ${\cal
S}_K^{fingen}(\Sigma)$ and ${\cal S}_K^{fin}(\Sigma)$.}
  
\end{figure}
 
\section{Multiplicity automata and rational stochastic languages.}\label{marsl}

In the previous Sections, we have defined several classes of rational
stochastic languages over $K\in \{{\mathbb R}, {\mathbb Q}, {\mathbb
R}^+, {\mathbb Q}^+\}$. In this section, we study the representation
of these classes by means of multiplicity automata: given a subclass
${\cal C}$ of rational stochastic languages over $K$, is there a
subset of $K$-multiplicity automata both simple to identify and
sufficient to generate the elements of ${\cal C}$? The first result we
prove is negative: it is undecidable whether a given multiplicity
automaton over ${\mathbb Q}$ generates a stochastic
language. Moreover, there exist no recursively enumerable subset of
multiplicity automata over ${\mathbb Q}$ sufficient to generate ${\cal
S}_{\mathbb Q}^{rat}(\Sigma)$. This result implies that no classes of
multiplicity automata can efficiently represent the class of rational
stochastic languages over ${\mathbb Q}$ or ${\mathbb R}$. In the other
hand, we show that the class of $K$-\emph{probabilistic automata}
represents ${\cal S}_K^{rat}(\Sigma)$ when $K\in \{{\mathbb R}^+,
{\mathbb Q}^+\}$. Clearly, it can be decided efficiently whether
a given multiplicity automaton is a probabilistic automaton. We show
also that the class of $K$-\emph{probabilistic residual automata}
represents the class ${\cal S}_K^{fingen}(\Sigma)$ for any $K\in
\{{\mathbb R}, {\mathbb R}^+, {\mathbb Q}, {\mathbb Q}^+\}$. We do not
know whether the class of \emph{probabilistic residual automata} is
decidable. However, we show that it contains a subclass which is decidable and sufficient to generate ${\cal
S}_K^{fingen}(\Sigma)$. Nevertheless, we show that deciding whether a
given MA is in this subclass is a PSPACE-complete. 
Finally, the class of \emph{probabilistic
deterministic automata} over ${\mathbb R}^+$ (resp. ${\mathbb Q}^+$),
which is clearly decidable, 
represents the class ${\cal S}_K^{fin}(\Sigma)$ when $K\in \{{\mathbb
R}, {\mathbb R}^+\}$ (resp. $K\in \{ {\mathbb Q}, {\mathbb Q}^+\}$).

To our knowledge, the decidability of the following problems is still open:  
\begin{itemize}
\item decide whether a given multiplicity automaton is
    equivalent to a probabilistic automaton, or a 
    probabilistic residual automaton or a probabilistic deterministic automaton;
  \item decide whether a given probabilistic automaton is
    equivalent to a
    probabilistic  residual automaton or a probabilistic deterministic automaton;
  \item decide whether a given probabilistic residual automaton is
    equivalent to a probabilistic deterministic automaton.
\end{itemize}

\subsection{The class of MA which generate stochastic languages
is undecidable}

A MA $A$ generates a stochastic language $p_A$ if and only if 
\begin{itemize}
\item $\forall w\in \Sigma^*, p_A(w)\geq 0$ and,
\item $\sum_{w\in\Sigma^*}p_A(w)=1$.
\end{itemize}

We first show that the second condition can be checked within
polynomial time.

We need the following result:

\begin{lemma}\cite{Gantmacher66,BlondelTsitsiklis00}\label{spectralradius}
Let $M$ be a square matrix with coefficients in ${\mathbb Q}$. It is
decidable within polynomial time whether $M^k$ converges to 0 when $k$
tends to infinity. 
\end{lemma}
\begin{proof}(Sketch)
  First, $M^k$ converges to 0 when $k$ tends to infinity if and only
  if the spectral radius $\rho(M)$ of $M$, i.e. the maximum of the
  magnitudes of its eigenvalues, satisfies $\rho(M)<1$.  
  
  Then, $M$ satisfies $\rho(M)<1$ iff the Lyapunov equation
$$MPM^t=P$$ has a positive-definite solution. In that case the
solution is unique. Since the Lyapunov equation is linear in the
unknown entries of $P$, we can compute a a solution $P$ in polynomial
time, or decide it does not exist. To check that $P$ is positive
definite, it is sufficient to compute the determinants of the
principal minors of $P$ and check that they are all positive. \qed
\end{proof}

\begin{proposition}\label{boundedMA}
  Let $A$ be an MA over ${\mathbb Q}$. It is decidable within
  polynomial time whether the sum $\sum_{k}P_A(\Sigma^k)$ converges.
  If the sum $P_A(\Sigma^*)=\sum_{k}P_A(\Sigma^k)$ converges, it can
  be computed within polynomial time.
\end{proposition}

\begin{proof}
  Let $A=\left\langle \Sigma,Q,\varphi,\iota,\tau\right\rangle $ where
  $Q=\{q_1, \ldots,q_n\}$ and
  let $M$ be the square matrix
  defined by
  $M[i,j]=\left[\varphi(q_i,\Sigma,q_j)\right]_{1\leq i,j\leq n}$. We have
  $P_A(\Sigma^k)=\iota_AM^k\tau_A$ where $\iota_A=(\iota(q_1),\ldots,\iota(q_n))$
  and $\tau_A=(\tau(q_1),\ldots,\tau(q_n))^t$.

  Let $E$ be the subspace of ${\mathbb R}^n$ spanned by $\{M^k\tau_A|k\in
  {\mathbb N}\}$ and let $F$ be a complementary subspace of $E$ in
  ${\mathbb R}^n$. Let $H=\{u\in E|\forall k\in {\mathbb N}, \ 
  \iota_AM^ku=0\}$. Clearly, $E$ and $H$ are stable under $M$.  Let $G$ be a
  complementary subspace of $H$ in $E$.  For any $u\in {\mathbb R}^n$,
  there exists a unique decomposition of the form $u=u_F+u_G+u_H$
  where $u_F\in F, u_G\in G$ and $u_H\in H$.  Let $p_F$, $p_H$ and
  $p_G$ be the projections on $F$, $G$ and $H$ defined by
  $p_F(u)=u_F$, $p_G(u)=u_g$ and $p_H(u)=u_H$. Let $P_F$, $P_H$ and
  $P_G$ be the corresponding matrices.

 First note that for any integer $k\geq 1$ and any $u\in E$, we have
  $P_GM^kP_Gu=(P_GMP_G)^ku$. This is clear when $k=1$. We have 
\begin{align*}
P_GM^{k+1}P_Gu&=P_GM^k(MP_Gu)\\&=P_GM^k[P_HMP_Gu+P_GMP_Gu]\textrm{
since }MP_Gu\in E\\
&=P_GM^kP_G[P_GMP_Gu]\textrm{ since }\forall v\in H, Mv\in H\textrm{ and }P_G(v)=0\\
&=(P_GMP_G)^{k+1}u\textrm{ from induction hypothesis}.
\end{align*}

Note also that for any integer $k$ and any $u\in E$,
\begin{align*}
\iota_AM^ku&=\iota_AM^k(P_Gu+P_Hu)\textrm{ since }u\in E\\
&=\iota_AM^kP_Gu\textrm{ since }\forall v\in H, Mv\in H\textrm{ and
}\iota_Av=0\\
&=\iota_A(P_GM^kP_Gu+P_HM^kP_Gu)\textrm{ since }M^kP_Gu\in E\\
&=\iota_AP_GM^kP_Gu\textrm{ since }\forall v\in H, \iota_Av=0\\
&=\iota_A(P_GMP_G)^ku.
\end{align*}


  We show now that $\sum_{k\in {\mathbb N}}\iota_AM^k\tau_A$ is convergent iff
  $\lim_{k\rightarrow \infty} (P_GMP_G)^k=0$.
  \begin{itemize}
  \item Suppose that $\lim_{k\rightarrow \infty} (P_GMP_G)^k=0$. Then
    $Id-P_GMP_G$ is inversible and $\sum_{k\in {\mathbb N}}(P_GMP_G)^k$
    converges to $(Id-P_GMP_G)^{-1}$. Therefore, $\sum_{k\in {\mathbb N}}\iota_AM^k\tau_A$
    converges to $\iota_A(Id-P_GMP_G)^{-1}\tau_A$.
  \item  Suppose now that $\sum_{k\in {\mathbb N}}\iota_AM^k\tau_A$ is
         convergent.

         There exists $\lambda>0$ such that for all $u\in G$, there
         exists $n\in {\mathbb N}$ such that $|\iota_AM^nu|\geq
         \lambda||u||$. Otherwise, there would exist a sequence $u_k$
         of elements of $G$ such that for all integer $n$,
         $|\iota_AM^n(u_k)|< ||u_k||/k$. Let $v_k=u_k/||u_k||$ and let
         $v_{\sigma(k)}$ a subsequence which converges to $v$. Check
         that we should have $||v||=1$, $v\in G$ and $\iota_AM^nv=0$
         for any integer $n$, which is impossible since $v\neq 0$.
         
         Let $\lambda$ satisfying this property. For any integers $m$
         and $k$, there exists $n_k$ such that
         $$|\iota_AM^{n_k}(P_GM^kP_G)(M^m\tau_A)|\geq \lambda
         ||(P_GM^kP_G)(M^m\tau_A)||=\lambda
         ||(P_GMP_G)^k(M^m\tau_A)||.$$

We have also 

\begin{align*}
\iota_AM^{n_k}(P_GM^kP_G)(M^m\tau_A)&=\iota_A(P_GMP_G)^{n_k}(P_GM^kP_G)(M^m\tau_A)\\
&=\iota_A(P_GMP_G)^{n_k+k}(M^m\tau_A)\\
&=\iota_AM^{n_k+k}(M^m\tau_A)\\
&=\iota_AM^{n_k+k+m}\tau_A.
\end{align*}

If we suppose that
$\iota_AM^k\tau_A\rightarrow 0$ when $k\rightarrow \infty$, we must have
$|(P_GM^kP_G)(M^m\tau_A)||\rightarrow 0$ when $k\rightarrow \infty$ for
any integer $m$. As $\{M^m\tau_A\}$ generates $E$, $P_GM^kP_G$ converges to 0. 
  \end{itemize}
  
  To sum up, $\sum_{k}P_A(\Sigma^k)$ is bounded iff $(P_GMP_G)^k$
  converges to 0, which is a polynomially decidable problem
  (Lemma~\ref{spectralradius}). 

When the sum $\sum_{k}P_A(\Sigma^k)$
  converges, it is equal to $\iota_A(Id-P_GMP_G)^{-1}\tau_A$ which
  can be computed within polynomial time.  \qed
\end{proof}

\begin{example}
  Consider the MA $A''$ described on Fig.~\ref{fig:nonFatou}.  We have
$$\iota_{A''}=(1,0), \tau_{A''}=(1/4,1/4)^t\textrm{ and }M=\left(
\begin{array}{ll}
\frac{3}{4}&0\\
0&\frac{3}{4}\\
\end{array}
\right)$$
We have $M\tau_{A''}=3/4\tau_{A''}$ and therefore, $E$ is the vector space spanned by $\tau_{A''}$. Let $F$ be the complementary space of $E$ spanned by the vector $(1,-1)^t$; we have $$H=\{0\}, G=E, P_G=\frac{1}{2}\left(
\begin{array}{ll}
1&1\\
1&1\\
\end{array}
\right), \textrm{ and }1-P_GMP_G=\frac{1}{8}\left(
\begin{array}{cc}
5&-3\\
-3&5\\
\end{array}
\right)$$
Check that the inverse of $1-P_GMP_G$ is equal to $$\frac{1}{2}\left(
\begin{array}{ll}
5&3\\
3&5\\
\end{array}
\right)$$
and that $\iota_A(Id-P_GMP_G)^{-1}\tau_A=1$. 
\end{example}

We prove now that it is undecidable whether a multiplicity over
${\mathbb Q}$ generates a stochastic language. In order to prove this
result, we use a reduction to a decision problem about \emph{acceptor
PAs}.  

An MA $\left\langle \Sigma,Q,\varphi,\iota,\tau\right\rangle $
is an \emph{acceptor PA} if
\begin{itemize}
\item $\varphi$, $\iota$ and $\tau$ are non negative
functions, 
\item $\sum_{q\in Q}\iota(q)=1$, 
\item $\forall q\in Q,\forall
x\in\Sigma,\sum_{r\in Q}\varphi(q,x,r)=1$ 
\item there exists a unique
terminal state $t$ and $\tau(t)=1$.
\end{itemize}

Blondel and Canterini have shown that given an acceptor PA $A$ over  ${\mathbb Q}$ and $\lambda\in {\mathbb Q}$, it is
undecidable whether there exists a word $w$ such that
$P_{A}(w)<\lambda$ (\cite{BlondelCanterini03}).

\begin{theorem}
It is undecidable whether an MA over ${\mathbb Q}$ generates a stochastic language.
\end{theorem}
\begin{proof}
For any rational series $r$ over $\Sigma$, let us denote by $\overline{r}$ the rational series defined by $$\overline{r}=\sum_{w\in \Sigma^*}\frac{r(w)}{(|\Sigma|+1)^{|w|+1}}.$$

  Let $A=\left\langle \Sigma,Q,\varphi,\iota,\tau\right\rangle$ be an
  acceptor PA over ${\mathbb Q}$ and let $\lambda\in
  {\mathbb Q}$. Let $B=\left\langle
  \Sigma,Q,\varphi_B,\iota,\tau_B\right\rangle$ be the MA defined by
  $\varphi_B(q,x,q')=\varphi(q,x,q')/(|\Sigma|+1)$ and
  $\tau_B(q)=\tau(q)/(|\Sigma|+1)$ for any states $q,q'\in Q$ and any
  $x\in \Sigma$.  Remark that $B$ is semi PA and that $r_B=\overline{r_A}$.
  
  The sum $s=\sum_{w\in \Sigma^*}r_B(w)$ is bounded by 1 from
  Prop.~\ref{PASto} and can be computed within polynomial time by
  using the Prop.~\ref{boundedMA}. Let $c_{\lambda}$ be the series
  defined by $c_{\lambda}(w)=\lambda$ for any word $w\in \Sigma^*$. 

\begin{itemize}
\item If $s< \lambda$, then there must exists a word w such that $P_A(w)<\lambda$ since $$\sum_{w\in \Sigma^*}\frac{\lambda}{(|\Sigma|+1)^{|w|+1}}=\lambda.$$
\item If $s=\lambda$, the rational series
  $1+\overline{r_A-c_{\lambda}}$ is a stochastic language iff $r_A(w)
  \geq \lambda$ for any word $w$.
\item If $s>\lambda$, the rational series $\frac{1}{s-\lambda}\cdot \overline{r_A-c_{\lambda}}$ is a stochastic language iff  $r_A(w) \geq
  \lambda$ for any word $w$.
\end{itemize}
Since in the two last cases, a multiplicity automaton which generates $1+\overline{r_A-c_{\lambda}}$ (resp.  $\frac{1}{s-\lambda}\cdot \overline{r_A-c_{\lambda}}$) can easily be derived from $A$, an algorithm able to decide whether an MA
generates a stochastic language could be used to solve the decision problem on PA acceptors.\qed
\end{proof}

A reduction to the following undecidable problem could have also been
used: it is undecidable whether a rational series over ${\mathbb Z}$
takes a negative value~\cite{SalomaaSoittola78}.

The set of multiplicity automata over ${\mathbb Q}$
which generate stochastic languages is not only not recursive: it
contains no recursively enumerable set able to generate ${\cal
  S}_{{\mathbb Q}}^{rat}(\Sigma)$.

\begin{theorem}No recursively
enumerable set of multiplicity automata over ${\mathbb Q}$ exactly generates ${\cal S}_{{\mathbb Q}}^{rat}(\Sigma)$.
\end{theorem}
\begin{proof}
  From Prop.~\ref{boundedMA}, the set ${\cal A}$ composed of the multiplicity automata $A$ over
  ${\mathbb Q}$ which satisfy $P_A(\Sigma^*)=1$ is recursively
  enumerable.

The subset ${\cal B}$ composed of the elements of ${\cal A}$ which
satisfy $$\exists \ w\in \Sigma^* P_A(w)<0$$
is recursively
  enumerable.

  Suppose that there exists a recursive enumeration $R_0, \ldots, R_n,
  \ldots$ of multiplicity automata over ${\mathbb Q}$ sufficient to
  generate ${\cal S}_{{\mathbb Q}}^{rat}(\Sigma)$ and let $w_0, \ldots, w_n,
  \ldots$ be an enumeration of $\Sigma^*$.

Consider the following algorithm:

\bigskip

\noindent \texttt{Input: a multiplicity automaton $A$ over ${\mathbb  Q}$}

\noindent \texttt{If $p_A(\Sigma^*)=1$ then}

\noindent \texttt{~~~~For $i\geq 0$ do}

\noindent \texttt{~~~~~~~~If $p_A(w_i)<0$ then output NO; exit; EndIf}

\noindent \texttt{~~~~~~~~If $A$ is equivalent to $R_i$ then output YES; exit; EndIf}

\noindent \texttt{~~~~EndFor}

\noindent \texttt{Else}

\noindent \texttt{~~~~output NO; exit}

\noindent \texttt{EndIf}

\bigskip

Since the equality $\sum_{w\in \Sigma^*}P_A(w)=1$ and the equivalence
of two multiplicity automata can be decided, this algorithm would end
on any input and decide whether $A$ generates a stochastic language.
Therefore, the enumeration $R_0, \ldots, R_n, \ldots$ cannot exist.
\qed
\end{proof}

\subsection{Probabilistic automata}

So, ${\cal S}_{{\mathbb Q}}^{rat}(\Sigma)$ and ${\cal S}_{{\mathbb R}}^{rat}(\Sigma)$ cannot be identified by any efficient subclass of multiplicity automata. In the other hand, ${\cal S}_{{\mathbb Q}^+}^{rat}(\Sigma)$ and ${\cal S}_{{\mathbb R}^+}^{rat}(\Sigma)$ can be described by probabilistic automata which form an easily identifiable subclass of multiplicity automata. 

\begin{proposition}\label{caracPA}
Let $K\in \{{\mathbb R}^+, {\mathbb Q}^+\}$ and let $p\in K\langle\langle \Sigma\rangle\rangle$. Then, $p$ is a stochastic language over $K$ iff there exists a $K$-probabilistic automaton $A$ such that  $p=r_A$. 
\end{proposition}

\begin{proof}
The only thing to prove is that if $p\in  {\cal S}_{K}^{rat}(\Sigma)$ then there exists a $K$-probabilistic automaton $A$ such that  $p=r_A$. 

From Theorem~\ref{caracS+}, there exist a finite subset $S$ of ${\cal
  S}_{K}^{rat}(\Sigma)$ which generates a stable subsemimodule of
$K\langle\langle \Sigma\rangle\rangle$ and such that $p\in
conv_{K}(S)$. Suppose that $S$ is minimal for inclusion. For any
$s,s'\in S$ and any $x\in \Sigma$, let $\alpha_s$ and
$\alpha_{s,s'}^x\in K$ such that $p=\sum_{s\in S}\alpha_ss$ and
$\dot{x}s=\sum_{s'\in S}\alpha_{s,s'}^xs'$.

Let $A=\left\langle
    \Sigma,S,\varphi,\iota,\tau\right\rangle $ be the MA defined by:
\begin{itemize}
\item $\iota(s)=\alpha_s$,
\item $\tau(s)=s(\varepsilon)$,
\item $\varphi(s,x,s')=\alpha_{s,s'}^x$
\end{itemize}
 for any $s,s'\in S$ and any $x\in \Sigma$.
From Claims 1 and 2, $p=r_A$.

Since $S\subseteq {\cal S}_{K}^{rat}(\Sigma)$, every state of $A$ is
co-accessible and since $S$ is minimal, every state of $A$ is
accessible. Therefore, $A$ is trimmed.

Note that $\sum_{s\in S}\iota(s)=\sum_{s\in S}\alpha_s=1$ since
elements of $\{p\}\cup S$ are stochastic languages. 
For any $s\in S$, 
\begin{align*} 
\tau(s)+\sum_{s'\in S,x\in
\Sigma}\varphi(s,x,s')&=s(\varepsilon)+\sum_{s'\in S,x\in
\Sigma}\alpha_{s,s'}^x\\
&=s(\varepsilon)+\sum_{x\in
\Sigma}\dot{x}s(\Sigma^*)\\
 &= s(\varepsilon)+\sum_{x\in \Sigma}s(x\Sigma^*)\\
&=1.
\end{align*} Then, $A$ is a PA.
\qed

\end{proof}

\subsection{Probabilistic residual automata}

For any $K\in \{{\mathbb R}^+, {\mathbb Q}^+\}$,
the class ${\cal S}_{K}^{fingen}(\Sigma)$ can be described by
probabilistic residual automata.

\begin{proposition}\label{caracPRA}
  Let $K\in \{{\mathbb R}^+, {\mathbb Q}^+\}$ and let $p\in
  K\langle\langle \Sigma\rangle\rangle$. Then, $p$ is a stochastic
  language over $K$ whose residual subsemimodule is finitely generated iff
  there exists a $K$-probabilistic residual automaton $A$ such
  that $p=r_A$.
\end{proposition}

\begin{proof}
\begin{itemize}
\item Let $p\in {\cal S}_{K}^{fingen}(\Sigma)$ and let $w_1, \ldots,
w_n\in res(p)$ be such that $S=\{w_1^{-1}p, \ldots, w_n^{-1}p\}$
generates $[Res(p)]$. Let $A$ be the MA
associated with $S$ as in the proof of Prop.~\ref{caracPA}. Check that
$A$ is a PRA which generates $p$.
\item Let $A\left\langle \Sigma,Q,\varphi,\iota,\tau\right\rangle $ be
  a PRA which generates $p$ and for any $q\in Q$, let $w_q\in
  \Sigma^*$ be such that $r_{A,q}=w_q^{-1}p$. From Claim~3,
  $\{w_q^{-1}p|q\in Q\}$ generates a stable subsemimodule $M$ which
  contains $p$. Check that $[Res(p)]=M$.
\end{itemize}\qed
\end{proof}

Remark that from Prop.~\ref{primeResidual}, there exists a unique
minimal subset $S$ of $Res(p)$ which generates  $[Res(p)]$. A PRA based on this set has a minimal number of
states.

We do not know whether the class of PRA is decidable. However, we show
that the class of ${\mathbb R}^+$-\emph{reduced PRA} is
decidable. Since a reduced PRA is a PRA, any PRA is equivalent to a
reduced PRA and therefore, this class is sufficient to generate ${\cal
S}_K^{fingen}(\Sigma)$.

Let $A$ be a PA and let $\left\langle \Sigma,Q,\delta,Q_I,
    Q_T\right\rangle$ be the support of $A$. If for any state $q\in Q$,
    there exists a word $w_q$ such that $\delta(Q_I,w_q)=\{q\}$, then
    $A$ is a PRA since $w_q^{-1}r_A=r_{A,q}$. The converse is true
    when $A$ is reduced.

\begin{proposition}\label{convexreducedMAandPRA}
  Let $A$ be a ${\mathbb R}^+$-reduced PA and let $\left\langle
    \Sigma,Q,\delta,Q_I, Q_T\right\rangle$ be the support of $A$. Then,
    $A$ is a PRA if and only if
  for any state $q\in Q$, there exists a word $w$ such that
  $\delta(Q_I,w)=\{q\}$.
\end{proposition}

\begin{proof}
Suppose that $A$ is a PRA. Let $q\in Q$ and $w$ be a word
such that $w_q^{-1}r_A=r_{A,q}$. Let $Q_w=\delta(Q_I,w)$. There exist
$(\alpha_{q'})_{q'\in Q_w}$ such that $w^{-1}r_A=\sum_{q'\in
Q_w}\alpha_{q'}r_{A,q'}$. Since $q\in Q_w$,
$(1-\alpha_q)r_{A,q}=\sum_{q'\in Q_w, q'\neq
q}\alpha_{q'}r_{A,q'}$. Since $A$ is ${\mathbb R}^+$-reduced, we must have
$\alpha_q=1$ and therefore, $Q_w=\{q\}$.\qed
\end{proof}

\begin{corollary}\label{decidePRA}
It can be decided whether
a ${\mathbb R}^+$-reduced MA is a PRA.
\end{corollary}

\begin{proof}
It can easily be decided whether an MA is a PA. Then, the power set
construction can be used to check whether any state can be uniquely
reached by some word. \qed
\end{proof}

From Prop.~\ref{Prop:Kreduced}, it can efficiently be decided whether
 an MA is ${\mathbb R}^+$-reduced PA.  But unfortunately, no \emph{efficient}
 decision procedure exist to decide whether it is an ${\mathbb
 R}^+$-reduced PRA: the decision problem is PSPACE-complete.

\begin{proposition}
Deciding whether a ${\mathbb R}^+$-reduced PA is a PRA is PSPACE-complete.
\end{proposition}

\begin{proof}
We prove the proposition by reduction of the following PSPACE-complete
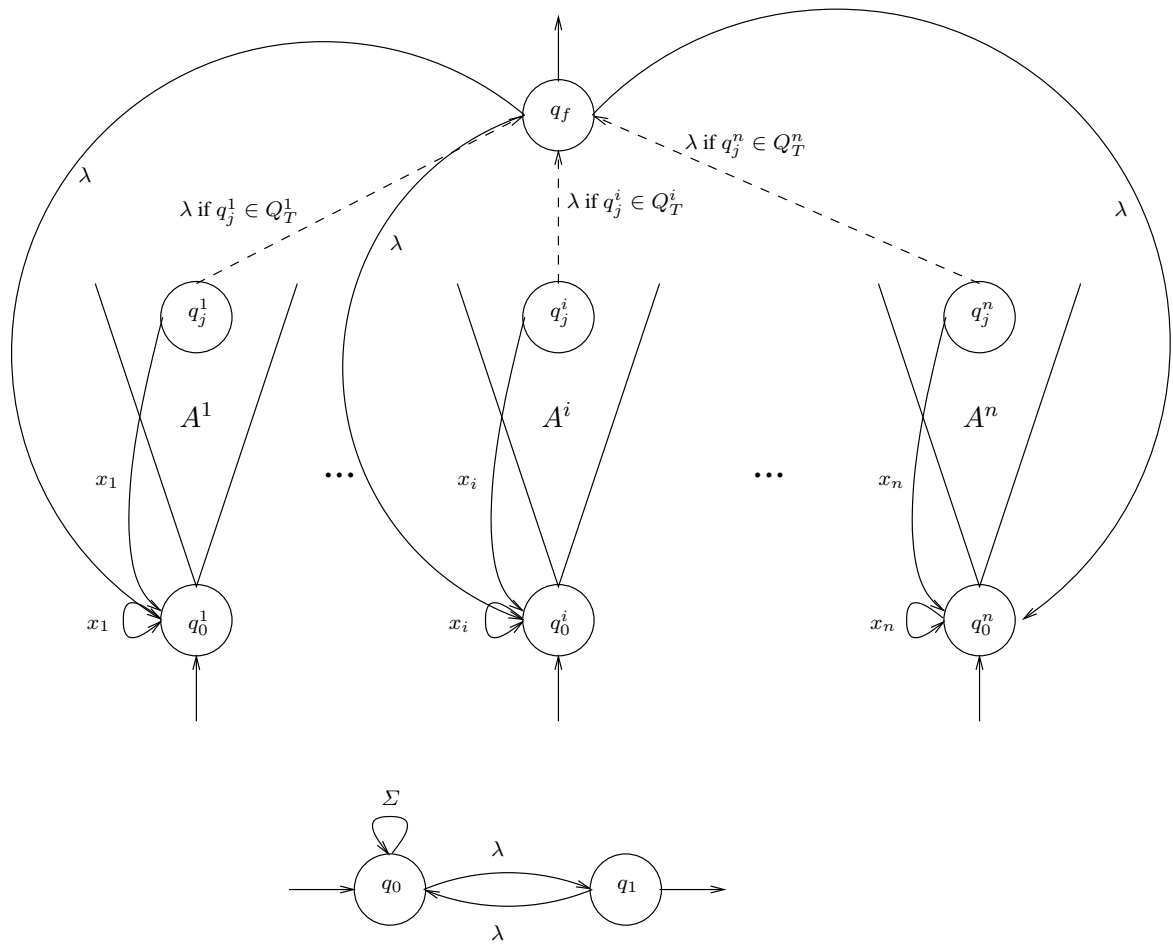
\begin{figure}[htbph]
  \centering
  \input{PraPspace.pstex_t}
  \caption{The union of the languages recognized by the automata $A_i$
  is different from $\Sigma^*$ if and only if this automaton is the
  support of a ${\mathbb R}^+$-reduced PRA.}
  
\end{figure}
problem: given $n$ DFA $A^1, \ldots, A^n$ over $\Sigma$, let $L_i$ be
the language recognized by $A^i$ for $1\leq i \leq n$, deciding
whether $\cup_{i=1}^n L_i=\Sigma^*$ is PSPACE-complete.

Let $A^i=\langle \Sigma, Q^i, \{q_0^i\}, Q_T^i, \delta^i\rangle$ for
$1\leq i \leq n$ where $i\neq j$ implies that $Q^i\cap Q^j=\emptyset$.
We may suppose that $L_i\neq \emptyset$ for $1\leq i \leq n$. Consider
3 new states $q_0, q_1, q_f$, $n+1$ new letters $x_1, \ldots, x_n,
\lambda$. Let $A=\langle \Sigma_A, Q_A, Q_I, Q_T, \delta\rangle$ be an
NFA defined by:
\begin{itemize}
\item $\Sigma_A=\Sigma\cup \{x_1, \ldots, x_n,
\lambda\}$
\item $Q_A=\cup_{i=1}^n Q^i \cup \{q_0, q_1, q_f\}$, 
\item $Q_I=\{q_0, q_0^1, \ldots, q_0^n\}$, 
\item $Q_T=\{q_1, q_f\}$, 
\item for any $1\leq i,j
\leq n$, any $q\in Q^i$ and any $x\in \Sigma$, 
\begin{itemize}
\item $\delta(q,x)=\delta^i(q,x)$, 
\item $\delta(q,x_j)=\{q_0^i\}$ if $i=j$ and $\emptyset$ otherwise, 
\item $\delta(q,\lambda)=\{q_f\}$ if $q\in Q_T^i$ and $\emptyset$ otherwise, 
\end{itemize}
\item for any $x\in \Sigma$, $\delta(q_0,x)=\{q_0\}$, $\delta(q_1,x)=\emptyset$ and $\delta(q_f,x)=\emptyset$,
\item $\delta(q_0,\lambda)=\{q_1\}$, $\delta(q_1,\lambda)=\{q_0\}$ and $\delta(q_f,\lambda)=\cup_{i=1}^n \{q_0^1, \ldots, q_0^n\}$.
\end{itemize}

Check that for any $q\in \cup_{i=1}^n Q^i \cup \{q_f\}$, there exists
a word $w_q$ such that $\delta(Q_I,w)=\{q\}$. If there exists a word
$w_0$ such that $\delta(Q_I,w_0)=\{q_0\}$ then
$\delta(Q_I,w_0\lambda)=\{q_1\}$.

Now, suppose that $\cup_{i=1}^n L_i\neq \Sigma^*$ and let $u\in
\Sigma^*\setminus \cup_{i=1}^n L_i$. Then $\delta(Q_I,u)\cap
\cup_{i=1}^n Q_T^i=\emptyset$ and therefore,
$\delta(Q_I,u\lambda)=\{q_1\}$ and
$\delta(Q_I,u\lambda\lambda)=\{q_0\}$.

If $\cup_{i=1}^n L_i= \Sigma^*$, for any $u\in \Sigma^*$,
$\delta(Q_I,u)\cap \cup_{i=1}^n
Q_T^i\neq \emptyset$,$\delta(Q_I,u\lambda)=\{q_1,q_f\},
\delta(Q_I,u\lambda\Sigma)=\emptyset$ and
$\delta(Q_I,u\lambda\lambda)=Q_I$. Therefore, there exists no word
$w_0$ such that $\delta(Q_I,w_0)=\{q_0\}$.

That is, $\cup_{i=1}^n L_i\neq \Sigma^*$ if and only if for any $q\in
Q_A$, there exists a word $w_q\in \Sigma_A^*$ such that $\delta(Q_I,w_q)=\{q\}$.

Now, associate a new letter $y_q$ to each state $q\in Q_A$ and consider the MA $B=\langle \Sigma_B, Q_B, \iota, \tau, \varphi\rangle$ where 
\begin{itemize}
\item $\Sigma_B=\Sigma_A\cup\{y_q|q\in Q_A\}$,
\item $Q_B=Q_A\cup\{q_b\}$,
\item $\iota(q)=1/(n+1)$ if $q\in Q_I$ and 0 otherwise,
\item $\tau(q)=1$ if $q=q_b$ and 0 otherwise,
\item $\varphi(q,x,q')=1/(\sum_{y\in \Sigma}|\delta(q,y)|+1)$ if
  $q,q'\in Q_A$, $x\in \Sigma_A$ and $q'\in \delta(q,x)$,
\item $\varphi(q,y_q,q_b)=1/(\sum_{y\in \Sigma}|\delta(q,y)|+1)$,
\item $\varphi(q,x,q')=0$ in all other cases. 
\end{itemize}
Check that $B$ is a PA. $B$ is ${\mathbb R}^+$-reduced since for any
$q\in Q_A$, $r_{B,q}(y_{q'})\neq 0$ iff $q=q'$ and
$r_{B,q}(\varepsilon)=0$. $B$ is a PRA if and only if for any $q\in
Q_A$, there exists a word $w_q\in \Sigma_A^*$ such that $\delta(Q_I,w_q)=\{q\}$.

Putting all together, we see that an algorithm which decides whether
$B$ is a PRA could be used to decide whether $\cup_{i=1}^n L_i\neq
\Sigma^*$.

As the problem is clearly PSPACE, it is PSPACE-complete. \qed  
\end{proof}

It has been shown in~\cite{DenisLemayTerlutte2002b} that for any
polynomial $p(\cdot)$, there exists an NFA $A=\langle \Sigma_A, Q,
Q_I, Q_T, \delta\rangle$ which satisfies the following properties:
\begin{itemize}
\item for any state $q$ of $A$, there exists a word $w\in \Sigma^*$ such that $\delta(Q_I,w)=\{q\}$,
\item for any state $q$ of $A$, all words $w$ which satisfy $\delta(Q_I,w)=\{q\}$ have a length greater than $p(|Q|)$.
\end{itemize}
These NFA are support of PRA which inherit of this property.

So, reduced PRA form a decidable family which is sufficient to
generate ${\cal S}_K^{fingen}(\Sigma)$ but the membership problem for
this family is not polynomial. We can restrict this family to obtain a
polynomially decidable family and still sufficient to generate ${\cal
  S}_K^{fingen}(\Sigma)$.

Let $A=\langle \Sigma, Q, \iota, \tau, \varphi\rangle$ be a PRA. $A$
is \emph{prefixial} if for any $q\in Q$, there exists $w_q\in \Sigma^*$
such that $w_q^{-1}r_A=r_{A,q}$ and such that $\{w_q|q\in Q\}$ is
prefixial.

It is polynomially decidable whether an MA is a prefixial PRA. 

Let $A=\langle \Sigma, Q, \iota, \tau, \varphi\rangle$ be a PRA, and
for any $q\in Q$, let $w_q\in \Sigma^*$ such that
$w_q^{-1}r_A=r_{A,q}$. Let $W=\{w_q|q\in Q\}$ and let $\overline{W}$
be the smallest prefixial subset of $\Sigma^*$ which contains $W$. Let
$B=\langle\Sigma, \overline{W}, \overline{\iota}, \overline{\tau},
\overline{\varphi}\rangle$ be the MA defined by:
\begin{itemize}
\item $\overline{\iota(q)}=1$ if $q=\varepsilon$ and 0 otherwise,
\item $\overline{\tau}(w)=w^{-1}r_A(\varepsilon)$,
\item $\overline{\varphi}(w,x,wx)=w^{-1}r_A(x\Sigma^*)$ for any $x\in \Sigma$,
\item $\overline{\varphi}(w_q,x,w_{q'})=\varphi(q,x,q')$ if $w_qx\not \in W$,
\item $\overline{\varphi}(w,x,w')=0$ in all other cases. 
\end{itemize}

It can be shown that $B$ is a prefixial PRA equivalent to $A$.

\subsection{Probabilistic Deterministic Automata}

For any $K\in \{{\mathbb R}, {\mathbb Q}, {\mathbb R}^+, {\mathbb Q}^+\}$,
the class ${\cal S}_{K}^{fin}(\Sigma)$ can be described by
probabilistic deterministic automata.

\begin{proposition}\label{caracPDA}
Let $K\in \{{\mathbb R}, {\mathbb Q}, {\mathbb R}^+, {\mathbb Q}^+\}$
and let $p\in K\langle\langle \Sigma\rangle\rangle$. Then, $p$ is a
stochastic language over $K$ which has finitely many residual languages iff there exists a $K$-probabilistic deterministic
automaton $A$ such that $p=r_A$.
\end{proposition}

\begin{proof}
From Prop~\ref{stofin}, we can suppose that $K\in \{{\mathbb R}^+, {\mathbb Q}^+\}$ . 

\begin{itemize}
\item Let $p\in {\cal S}_{K}^{fin}(\Sigma)$ and let
$Res(p)=\{w_1^{-1}p, \ldots, w_n^{-1}p\}$. Let $A$ be the MA
associated with $S$ as in the proof of Prop~\ref{caracPA}. As there
exists $i\in \{1, \ldots, n\}$ such that $p=w_i^{-1}p$, we can suppose
that $\alpha_s=1$ if $s= w_i^{-1}p$ and 0 otherwise. Let $sw_i^{-1}p$. If
$x\not\in res(s)$, then $\sum_{w\in \Sigma^*}p(w_ixw)=0$ and
since $K\in \{{\mathbb R}^+, {\mathbb Q}^+\}$, this implies that
$p(w_ixw)=0$ for any word $w$. Therefore, in this case, it is possible
to choose $\alpha_{s,s'}^x=0$ for any $s'\in Res(p)$. When $x\in res(s)$, there exists $j\in \{1, \ldots, n\}$ such that $x^{-1}s=w_j^{-1}p$. In this case, we can choose $\alpha_{s,s'}^x=1$ if $s'= w_j^{-1}p$ and 0 otherwise.

Then, check that
$A$ is a PDA which generates $p$.
\item Let $A=\left\langle
    \Sigma,Q,\varphi,\iota,\tau\right\rangle $ be a PDA which generates $p$ and let $Q_I=\{q_0\}$. For any $w\in \Sigma^*$, there eixts only one state $q\in Q$ such that $\varphi(q_0,w,q)\neq 0$. Therefore, $Res(p)\subseteq \{r_{A,q}|q\in Q\}$ and $Res(p)$ is a finite state. 
\end{itemize}\qed
\end{proof}

\begin{figure}[htbp]\label{recapitulatif3}
  \input{recapitulatif3.pstex_t}
  \caption{Inclusion relations between classes of classes of rational
  stochastic languages.}
\end{figure}
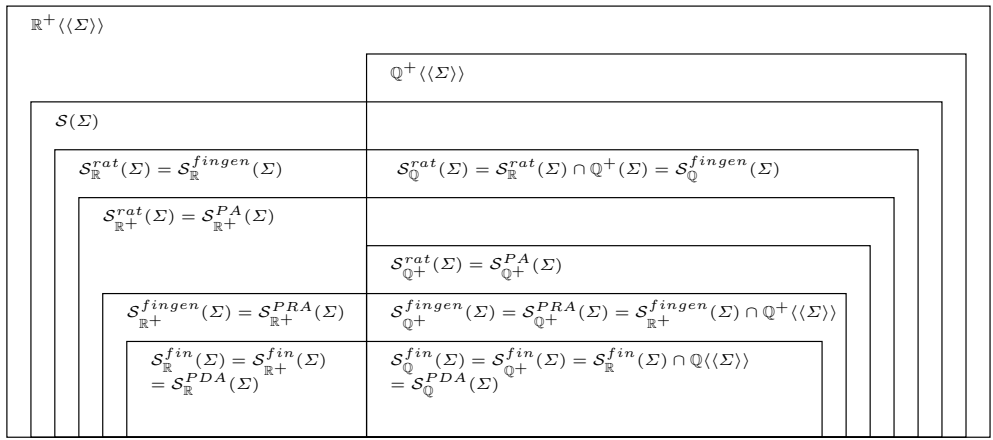

\section{Conclusion}

In this paper, we have carried out a systematic study of
rational stochastic languages, which are precisely the objects
probabilistic grammatical inference deal with. This study, and the
results we bring out, whether they are original or derived from
former contributions, support our opinion that researches in grammatical
inference should be based and rely on formal language theory. Doing
this makes it possible to reuse powerful tools and general results for
inference purposes. Moreover, this approach may help finding
out what particular properties are important for grammatical
inference. For example, a learning sample $\{w_1, \ldots, w_n\}$
independently drawn according to a target stochastic language $p$
provides statistical information on the residual languages of $p$. In
order to infer an approximation of $p$ by means of a multiplicity automata
$A$, there should be a structural link between the states of $A$ and
the observed data and hence, between the states of $A$ and the
residual languages of $p$. This explains why most results in
grammatical inference deal with PDA and PRA, i.e. classes of
multiplicity automata for which there exists a strong connection
between the states and the residual languages of the stochastic
languages they generate. This also explains why there is no useful
general inference result about PA: the residual subsemimodule of
a rational stochastic language over ${\mathbb R}^+$ or ${\mathbb Q}^+$
may be not finitely generated and hence, no finite set of residual
languages can be used to represent it. Moreover, PA admits no natural
normal form. On the other hand, the residual subsemimodule of rational
stochastic languages over ${\mathbb R}$ or ${\mathbb Q}$ are finitely
generated and admit a basis made of residual languages. Even if
there exists no recursively enumerable subset of MA capable of
generating them, this study has encouraged us to try to find a way to infer
these most general stochastic languages.
See~\cite{DenisEspositoHabrardTR06} for preliminary results. We are
also currently working on \emph{tree rational stochastic languages},
following a similar approach, in order to deal with tree probabilistic
languages inference. This work is still in progress.


\end{document}

%% file: nonfingen.pstex_t
\begin{picture}(0,0)%
\includegraphics{nonfingen.pstex}%
\end{picture}%
\setlength{\unitlength}{3947sp}%
\begingroup\makeatletter\ifx\SetFigFont\undefined%
\gdef\SetFigFont#1#2#3#4#5{%
  \reset@font\fontsize{#1}{#2pt}%
  \fontfamily{#3}\fontseries{#4}\fontshape{#5}%
  \selectfont}%
\fi\endgroup%
\begin{picture}(4973,3125)(289,-3774)
\put(2738,-2536){\makebox(0,0)[lb]{\smash{{\SetFigFont{6}{7.2}{\rmdefault}{\mddefault}{\updefault}{\color[rgb]{0,0,0}$p_1$}%
}}}}
\put(3938,-2536){\makebox(0,0)[lb]{\smash{{\SetFigFont{6}{7.2}{\rmdefault}{\mddefault}{\updefault}{\color[rgb]{0,0,0}$p_2$}%
}}}}
\put(3488,-2536){\makebox(0,0)[lb]{\smash{{\SetFigFont{6}{7.2}{\rmdefault}{\mddefault}{\updefault}{\color[rgb]{0,0,0}$p$}%
}}}}
\put(3413,-3774){\makebox(0,0)[lb]{\smash{{\SetFigFont{6}{7.2}{\rmdefault}{\mddefault}{\updefault}{\color[rgb]{0,0,0}O}%
}}}}
\put(488,-2536){\makebox(0,0)[lb]{\smash{{\SetFigFont{6}{7.2}{\rmdefault}{\mddefault}{\updefault}{\color[rgb]{0,0,0}$p_3$}%
}}}}
\put(3488,-3249){\makebox(0,0)[lb]{\smash{{\SetFigFont{6}{7.2}{\rmdefault}{\mddefault}{\updefault}{\color[rgb]{0,0,0}$\dot{a}p$}%
}}}}
\put(3488,-3474){\makebox(0,0)[lb]{\smash{{\SetFigFont{6}{7.2}{\rmdefault}{\mddefault}{\updefault}{\color[rgb]{0,0,0}$\dot{\overline{a^2}}p$}%
}}}}
\put(451,-736){\makebox(0,0)[lb]{\smash{{\SetFigFont{6}{7.2}{\rmdefault}{\mddefault}{\updefault}{\color[rgb]{0,0,0}${\cal V}$}%
}}}}
\put(2551,-774){\makebox(0,0)[lb]{\smash{{\SetFigFont{6}{7.2}{\rmdefault}{\mddefault}{\updefault}{\color[rgb]{0,0,0}${\cal V}_p$}%
}}}}
\put(1013,-2574){\makebox(0,0)[lb]{\smash{{\SetFigFont{6}{7.2}{\rmdefault}{\mddefault}{\updefault}{\color[rgb]{0,0,0}${\cal S}(\Sigma)\cap{\cal V}$}%
}}}}
\put(3525,-2311){\makebox(0,0)[lb]{\smash{{\SetFigFont{6}{7.2}{\rmdefault}{\mddefault}{\updefault}{\color[rgb]{0,0,0}$a^{-1}p$}%
}}}}
\put(2326,-2311){\makebox(0,0)[lb]{\smash{{\SetFigFont{6}{7.2}{\rmdefault}{\mddefault}{\updefault}{\color[rgb]{0,0,0}$(a^2)^{-1}p$}%
}}}}
\end{picture}%

%% file: recapitulatif1.pstex_t
\begin{picture}(0,0)%
\includegraphics{recapitulatif1.pstex}%
\end{picture}%
\setlength{\unitlength}{3947sp}%
\begingroup\makeatletter\ifx\SetFigFont\undefined%
\gdef\SetFigFont#1#2#3#4#5{%
  \reset@font\fontsize{#1}{#2pt}%
  \fontfamily{#3}\fontseries{#4}\fontshape{#5}%
  \selectfont}%
\fi\endgroup%
\begin{picture}(6174,2724)(589,-2173)
\put(751,389){\makebox(0,0)[lb]{\smash{{\SetFigFont{6}{7.2}{\rmdefault}{\mddefault}{\updefault}{\color[rgb]{0,0,0}${\mathbb R^+}\langle\langle\Sigma\rangle\rangle$}%
}}}}
\put(3001, 89){\makebox(0,0)[lb]{\smash{{\SetFigFont{6}{7.2}{\rmdefault}{\mddefault}{\updefault}{\color[rgb]{0,0,0}${\mathbb Q^+}\langle\langle\Sigma\rangle\rangle$}%
}}}}
\put(901,-211){\makebox(0,0)[lb]{\smash{{\SetFigFont{6}{7.2}{\rmdefault}{\mddefault}{\updefault}{\color[rgb]{0,0,0}${\cal S}(\Sigma)$}%
}}}}
\put(3001,-211){\makebox(0,0)[lb]{\smash{{\SetFigFont{6}{7.2}{\rmdefault}{\mddefault}{\updefault}{\color[rgb]{0,0,0}${\cal S}(\Sigma)\cap{\mathbb Q^+}\langle\langle\Sigma\rangle\rangle$}%
}}}}
\put(1201,-811){\makebox(0,0)[lb]{\smash{{\SetFigFont{6}{7.2}{\rmdefault}{\mddefault}{\updefault}{\color[rgb]{0,0,0}${\cal S}^{rat}_{\mathbb R^+}(\Sigma)$}%
}}}}
\put(3001,-1111){\makebox(0,0)[lb]{\smash{{\SetFigFont{6}{7.2}{\rmdefault}{\mddefault}{\updefault}{\color[rgb]{0,0,0}${\cal S}^{rat}_{\mathbb Q^+}(\Sigma)$}%
}}}}
\put(1051,-511){\makebox(0,0)[lb]{\smash{{\SetFigFont{6}{7.2}{\rmdefault}{\mddefault}{\updefault}{\color[rgb]{0,0,0}${\cal S}^{rat}_{\mathbb R}(\Sigma)$}%
}}}}
\put(3001,-811){\makebox(0,0)[lb]{\smash{{\SetFigFont{6}{7.2}{\rmdefault}{\mddefault}{\updefault}{\color[rgb]{0,0,0}${\cal S}^{rat}_{\mathbb R^+}(\Sigma)\cap{\mathbb Q^+}\langle\langle\Sigma\rangle\rangle$}%
}}}}
\put(3039,-511){\makebox(0,0)[lb]{\smash{{\SetFigFont{6}{7.2}{\rmdefault}{\mddefault}{\updefault}{\color[rgb]{0,0,0}${\cal S}^{rat}_{\mathbb Q}(\Sigma)={\cal S}^{rat}_{\mathbb R}(\Sigma)\cap {\mathbb Q}^+(\Sigma)$}%
}}}}
\end{picture}%

%% file: recapitulatif2.pstex_t
\begin{picture}(0,0)%
\includegraphics{recapitulatif3.pstex}%
\end{picture}%
\setlength{\unitlength}{3947sp}%
\begingroup\makeatletter\ifx\SetFigFont\undefined%
\gdef\SetFigFont#1#2#3#4#5{%
  \reset@font\fontsize{#1}{#2pt}%
  \fontfamily{#3}\fontseries{#4}\fontshape{#5}%
  \selectfont}%
\fi\endgroup%
\begin{picture}(6174,2723)(589,-2172)
\put(3039,-511){\makebox(0,0)[lb]{\smash{{\SetFigFont{6}{7.2}{\rmdefault}{\mddefault}{\updefault}{\color[rgb]{0,0,0}${\cal S}^{rat}_{\mathbb Q}(\Sigma)={\cal S}^{rat}_{\mathbb R}(\Sigma)\cap {\mathbb Q}^+(\Sigma)={\cal S}^{fingen}_{\mathbb Q}(\Sigma)$}%
}}}}
\put(3001,-1410){\makebox(0,0)[lb]{\smash{{\SetFigFont{6}{7.2}{\rmdefault}{\mddefault}{\updefault}{\color[rgb]{0,0,0}${\cal S}^{fingen}_{\mathbb Q^+}(\Sigma)={\cal S}^{fingen}_{\mathbb R^+}(\Sigma)\cap{\mathbb Q^+}\langle\langle\Sigma\rangle\rangle$}%
}}}}
\put(3001,-1710){\makebox(0,0)[lb]{\smash{{\SetFigFont{6}{7.2}{\rmdefault}{\mddefault}{\updefault}{\color[rgb]{0,0,0}${\cal S}^{fin}_{\mathbb Q}(\Sigma)={\cal S}^{fin}_{\mathbb Q^+}(\Sigma)={\cal S}^{fin}_{\mathbb R}(\Sigma)\cap{\mathbb Q}\langle\langle\Sigma\rangle\rangle$}%
}}}}
\put(751,389){\makebox(0,0)[lb]{\smash{{\SetFigFont{6}{7.2}{\rmdefault}{\mddefault}{\updefault}{\color[rgb]{0,0,0}${\mathbb R^+}\langle\langle\Sigma\rangle\rangle$}%
}}}}
\put(3001, 89){\makebox(0,0)[lb]{\smash{{\SetFigFont{6}{7.2}{\rmdefault}{\mddefault}{\updefault}{\color[rgb]{0,0,0}${\mathbb Q^+}\langle\langle\Sigma\rangle\rangle$}%
}}}}
\put(3001,-1110){\makebox(0,0)[lb]{\smash{{\SetFigFont{6}{7.2}{\rmdefault}{\mddefault}{\updefault}{\color[rgb]{0,0,0}${\cal S}^{rat}_{\mathbb Q^+}(\Sigma)$}%
}}}}
\put(901,-211){\makebox(0,0)[lb]{\smash{{\SetFigFont{6}{7.2}{\rmdefault}{\mddefault}{\updefault}{\color[rgb]{0,0,0}${\cal S}(\Sigma)$}%
}}}}
\put(1051,-511){\makebox(0,0)[lb]{\smash{{\SetFigFont{6}{7.2}{\rmdefault}{\mddefault}{\updefault}{\color[rgb]{0,0,0}${\cal S}^{rat}_{\mathbb R}(\Sigma)={\cal S}^{fingen}_{\mathbb R}(\Sigma)$}%
}}}}
\put(1201,-810){\makebox(0,0)[lb]{\smash{{\SetFigFont{6}{7.2}{\rmdefault}{\mddefault}{\updefault}{\color[rgb]{0,0,0}${\cal S}^{rat}_{\mathbb R^+}(\Sigma)$}%
}}}}
\put(1351,-1410){\makebox(0,0)[lb]{\smash{{\SetFigFont{6}{7.2}{\rmdefault}{\mddefault}{\updefault}{\color[rgb]{0,0,0}${\cal S}^{fingen}_{\mathbb R^+}(\Sigma)$}%
}}}}
\put(1501,-1710){\makebox(0,0)[lb]{\smash{{\SetFigFont{6}{7.2}{\rmdefault}{\mddefault}{\updefault}{\color[rgb]{0,0,0}${\cal S}^{fin}_{\mathbb R}(\Sigma)={\cal S}^{fin}_{\mathbb R^+}(\Sigma)$}%
}}}}
\end{picture}%

%% file: PraPspace.pstex_t
\begin{picture}(0,0)%
\epsfig{file=PraPspace.pstex}%
\end{picture}%
\setlength{\unitlength}{3947sp}%
\begingroup\makeatletter\ifx\SetFigFont\undefined%
\gdef\SetFigFont#1#2#3#4#5{%
  \reset@font\fontsize{#1}{#2pt}%
  \fontfamily{#3}\fontseries{#4}\fontshape{#5}%
  \selectfont}%
\fi\endgroup%
\begin{picture}(7295,5895)(1098,-5293)
\put(2162,-709){\makebox(0,0)[lb]{\smash{{\SetFigFont{8}{9.6}{\rmdefault}{\mddefault}{\updefault}{\color[rgb]{0,0,0}$\lambda$ if $q_j^1\in Q^1_T$}%
}}}}
\put(3854,-3301){\makebox(0,0)[lb]{\smash{{\SetFigFont{8}{9.6}{\rmdefault}{\mddefault}{\updefault}{\color[rgb]{0,0,0}$x_i$}%
}}}}
\put(3907,-2401){\makebox(0,0)[lb]{\smash{{\SetFigFont{8}{9.6}{\rmdefault}{\mddefault}{\updefault}{\color[rgb]{0,0,0}$x_i$}%
}}}}
\put(4436,-2031){\makebox(0,0)[lb]{\smash{{\SetFigFont{11}{13.2}{\rmdefault}{\mddefault}{\updefault}{\color[rgb]{0,0,0}$A^i$}%
}}}}
\put(4489,-3301){\makebox(0,0)[lb]{\smash{{\SetFigFont{8}{9.6}{\rmdefault}{\mddefault}{\updefault}{\color[rgb]{0,0,0}$q_0^i$}%
}}}}
\put(2162,-2031){\makebox(0,0)[lb]{\smash{{\SetFigFont{11}{13.2}{\rmdefault}{\mddefault}{\updefault}{\color[rgb]{0,0,0}$A^1$}%
}}}}
\put(1633,-2401){\makebox(0,0)[lb]{\smash{{\SetFigFont{8}{9.6}{\rmdefault}{\mddefault}{\updefault}{\color[rgb]{0,0,0}$x_1$}%
}}}}
\put(1580,-3301){\makebox(0,0)[lb]{\smash{{\SetFigFont{8}{9.6}{\rmdefault}{\mddefault}{\updefault}{\color[rgb]{0,0,0}$x_1$}%
}}}}
\put(2215,-3301){\makebox(0,0)[lb]{\smash{{\SetFigFont{8}{9.6}{\rmdefault}{\mddefault}{\updefault}{\color[rgb]{0,0,0}$q_0^1$}%
}}}}
\put(7081,-2031){\makebox(0,0)[lb]{\smash{{\SetFigFont{11}{13.2}{\rmdefault}{\mddefault}{\updefault}{\color[rgb]{0,0,0}$A^n$}%
}}}}
\put(6552,-2401){\makebox(0,0)[lb]{\smash{{\SetFigFont{8}{9.6}{\rmdefault}{\mddefault}{\updefault}{\color[rgb]{0,0,0}$x_n$}%
}}}}
\put(6499,-3301){\makebox(0,0)[lb]{\smash{{\SetFigFont{8}{9.6}{\rmdefault}{\mddefault}{\updefault}{\color[rgb]{0,0,0}$x_n$}%
}}}}
\put(7134,-3301){\makebox(0,0)[lb]{\smash{{\SetFigFont{8}{9.6}{\rmdefault}{\mddefault}{\updefault}{\color[rgb]{0,0,0}$q_0^n$}%
}}}}
\put(3061,-2349){\makebox(0,0)[lb]{\smash{{\SetFigFont{17}{20.4}{\rmdefault}{\mddefault}{\updefault}{\color[rgb]{0,0,0}...}%
}}}}
\put(5759,-2349){\makebox(0,0)[lb]{\smash{{\SetFigFont{17}{20.4}{\rmdefault}{\mddefault}{\updefault}{\color[rgb]{0,0,0}...}%
}}}}
\put(4595,-656){\makebox(0,0)[lb]{\smash{{\SetFigFont{8}{9.6}{\rmdefault}{\mddefault}{\updefault}{\color[rgb]{0,0,0}$\lambda$ if $q_j^i\in Q^i_T$}%
}}}}
\put(3431,-4412){\makebox(0,0)[lb]{\smash{{\SetFigFont{8}{9.6}{\rmdefault}{\mddefault}{\updefault}{\color[rgb]{0,0,0}$\Sigma$}%
}}}}
\put(4119,-4729){\makebox(0,0)[lb]{\smash{{\SetFigFont{8}{9.6}{\rmdefault}{\mddefault}{\updefault}{\color[rgb]{0,0,0}$\lambda$}%
}}}}
\put(4119,-5258){\makebox(0,0)[lb]{\smash{{\SetFigFont{8}{9.6}{\rmdefault}{\mddefault}{\updefault}{\color[rgb]{0,0,0}$\lambda$}%
}}}}
\put(4489,-74){\makebox(0,0)[lb]{\smash{{\SetFigFont{8}{9.6}{\rmdefault}{\mddefault}{\updefault}{\color[rgb]{0,0,0}$q_f$}%
}}}}
\put(3431,-4941){\makebox(0,0)[lb]{\smash{{\SetFigFont{8}{9.6}{\rmdefault}{\mddefault}{\updefault}{\color[rgb]{0,0,0}$q_0$}%
}}}}
\put(4912,-4941){\makebox(0,0)[lb]{\smash{{\SetFigFont{8}{9.6}{\rmdefault}{\mddefault}{\updefault}{\color[rgb]{0,0,0}$q_1$}%
}}}}
\put(1527,-497){\makebox(0,0)[lb]{\smash{{\SetFigFont{8}{9.6}{\rmdefault}{\mddefault}{\updefault}{\color[rgb]{0,0,0}$\lambda$}%
}}}}
\put(3484,-920){\makebox(0,0)[lb]{\smash{{\SetFigFont{8}{9.6}{\rmdefault}{\mddefault}{\updefault}{\color[rgb]{0,0,0}$\lambda$}%
}}}}
\put(4489,-1343){\makebox(0,0)[lb]{\smash{{\SetFigFont{8}{9.6}{\rmdefault}{\mddefault}{\updefault}{\color[rgb]{0,0,0}$q_j^i$}%
}}}}
\put(7134,-1343){\makebox(0,0)[lb]{\smash{{\SetFigFont{8}{9.6}{\rmdefault}{\mddefault}{\updefault}{\color[rgb]{0,0,0}$q_j^n$}%
}}}}
\put(2215,-1343){\makebox(0,0)[lb]{\smash{{\SetFigFont{8}{9.6}{\rmdefault}{\mddefault}{\updefault}{\color[rgb]{0,0,0}$q_j^1$}%
}}}}
\put(8033,-709){\makebox(0,0)[lb]{\smash{{\SetFigFont{8}{9.6}{\rmdefault}{\mddefault}{\updefault}{\color[rgb]{0,0,0}$\lambda$}%
}}}}
\put(5336,-285){\makebox(0,0)[lb]{\smash{{\SetFigFont{8}{9.6}{\rmdefault}{\mddefault}{\updefault}{\color[rgb]{0,0,0}$\lambda$ if $q_j^n\in Q^n_T$}%
}}}}
\end{picture}%

%% file: recapitulatif3.pstex_t
\begin{picture}(0,0)%
\includegraphics{recapitulatif3.pstex}%
\end{picture}%
\setlength{\unitlength}{3947sp}%
\begingroup\makeatletter\ifx\SetFigFont\undefined%
\gdef\SetFigFont#1#2#3#4#5{%
  \reset@font\fontsize{#1}{#2pt}%
  \fontfamily{#3}\fontseries{#4}\fontshape{#5}%
  \selectfont}%
\fi\endgroup%
\begin{picture}(6199,2734)(589,-2183)
\put(3048,-515){\makebox(0,0)[lb]{\smash{{\SetFigFont{6}{7.2}{\rmdefault}{\mddefault}{\updefault}{\color[rgb]{0,0,0}${\cal S}^{rat}_{\mathbb Q}(\Sigma)={\cal S}^{rat}_{\mathbb R}(\Sigma)\cap {\mathbb Q}^+(\Sigma)={\cal S}^{fingen}_{\mathbb Q}(\Sigma)$}%
}}}}
\put(3011,-1719){\makebox(0,0)[lb]{\smash{{\SetFigFont{6}{7.2}{\rmdefault}{\mddefault}{\updefault}{\color[rgb]{0,0,0}${\cal S}^{fin}_{\mathbb Q}(\Sigma)={\cal S}^{fin}_{\mathbb Q^+}(\Sigma)={\cal S}^{fin}_{\mathbb R}(\Sigma)\cap{\mathbb Q}\langle\langle\Sigma\rangle\rangle$}%
}}}}
\put(752,388){\makebox(0,0)[lb]{\smash{{\SetFigFont{6}{7.2}{\rmdefault}{\mddefault}{\updefault}{\color[rgb]{0,0,0}${\mathbb R^+}\langle\langle\Sigma\rangle\rangle$}%
}}}}
\put(3011, 87){\makebox(0,0)[lb]{\smash{{\SetFigFont{6}{7.2}{\rmdefault}{\mddefault}{\updefault}{\color[rgb]{0,0,0}${\mathbb Q^+}\langle\langle\Sigma\rangle\rangle$}%
}}}}
\put(902,-214){\makebox(0,0)[lb]{\smash{{\SetFigFont{6}{7.2}{\rmdefault}{\mddefault}{\updefault}{\color[rgb]{0,0,0}${\cal S}(\Sigma)$}%
}}}}
\put(1053,-515){\makebox(0,0)[lb]{\smash{{\SetFigFont{6}{7.2}{\rmdefault}{\mddefault}{\updefault}{\color[rgb]{0,0,0}${\cal S}^{rat}_{\mathbb R}(\Sigma)={\cal S}^{fingen}_{\mathbb R}(\Sigma)$}%
}}}}
\put(1505,-1719){\makebox(0,0)[lb]{\smash{{\SetFigFont{6}{7.2}{\rmdefault}{\mddefault}{\updefault}{\color[rgb]{0,0,0}${\cal S}^{fin}_{\mathbb R}(\Sigma)={\cal S}^{fin}_{\mathbb R^+}(\Sigma)$}%
}}}}
\put(1203,-816){\makebox(0,0)[lb]{\smash{{\SetFigFont{6}{7.2}{\rmdefault}{\mddefault}{\updefault}{\color[rgb]{0,0,0}${\cal S}^{rat}_{\mathbb R^+}(\Sigma)={\cal S}^{PA}_{\mathbb R^+}(\Sigma)$}%
}}}}
\put(3011,-1117){\makebox(0,0)[lb]{\smash{{\SetFigFont{6}{7.2}{\rmdefault}{\mddefault}{\updefault}{\color[rgb]{0,0,0}${\cal S}^{rat}_{\mathbb Q^+}(\Sigma)={\cal S}^{PA}_{\mathbb Q^+}(\Sigma)$}%
}}}}
\put(1354,-1418){\makebox(0,0)[lb]{\smash{{\SetFigFont{6}{7.2}{\rmdefault}{\mddefault}{\updefault}{\color[rgb]{0,0,0}${\cal S}^{fingen}_{\mathbb R^+}(\Sigma)={\cal S}^{PRA}_{\mathbb R^+}(\Sigma)$}%
}}}}
\put(3011,-1418){\makebox(0,0)[lb]{\smash{{\SetFigFont{6}{7.2}{\rmdefault}{\mddefault}{\updefault}{\color[rgb]{0,0,0}${\cal S}^{fingen}_{\mathbb Q^+}(\Sigma)={\cal S}^{PRA}_{\mathbb Q^+}(\Sigma)={\cal S}^{fingen}_{\mathbb R^+}(\Sigma)\cap{\mathbb Q^+}\langle\langle\Sigma\rangle\rangle$}%
}}}}
\put(1505,-1870){\makebox(0,0)[lb]{\smash{{\SetFigFont{6}{7.2}{\rmdefault}{\mddefault}{\updefault}{\color[rgb]{0,0,0}$={\cal S}^{PDA}_{\mathbb R}(\Sigma)$}%
}}}}
\put(3011,-1870){\makebox(0,0)[lb]{\smash{{\SetFigFont{6}{7.2}{\rmdefault}{\mddefault}{\updefault}{\color[rgb]{0,0,0}$={\cal S}^{PDA}_{\mathbb Q}(\Sigma)$}%
}}}}
\end{picture}%

%% file: pra4.bbl
\begin{thebibliography}{ELDD02}

\bibitem[AW92]{AbeWarmuth92}
N.~Abe and M.~Warmuth.
\newblock On the computational complexity of approximating distributions by
  probabilistic automata.
\newblock {\em Machine Learning}, 9:205--260, 1992.

\bibitem[BC03]{BlondelCanterini03}
V.~D. Blondel and V.~Canterini.
\newblock Undecidable problems for probabilistic automata of fixed dimension.
\newblock {\em Theory of Computing Systems}, 36(3):231--245, 2003.

\bibitem[BR84]{BerstelReutenauer84}
J.~Berstel and C.~Reutenauer.
\newblock {\em Les s\'eries rationnelles et leurs langages}.
\newblock Masson, 1984.

\bibitem[BT00]{BlondelTsitsiklis00}
V.~D. Blondel and J.~N. Tsitsiklis.
\newblock A survey of computational complexity results in systems and control.
\newblock {\em Automatica}, 36(9):1249--1274, September 2000.

\bibitem[CO94]{CarrascoOncina94}
R.C. Carrasco and J.~Oncina.
\newblock Learning stochastic regular grammars by means of a state merging
  method.
\newblock In {\em ICGI}, pages 139--152, Heidelberg, September 1994.
  Springer-Verlag.

\bibitem[CO99]{CarrascoOncina99}
R.~C. Carrasco and J.~Oncina.
\newblock Learning deterministic regular grammars from stochastic samples in
  polynomial time.
\newblock {\em RAI}, (1):1--20, 1999.

\bibitem[DDE05]{DupontDenisEsposito05}
P.~Dupont, F.~Denis, and Y.~Esposito.
\newblock Links between probabilistic automata and hidden markov models:
  probability distributions, learning models and induction algorithms.
\newblock {\em Pattern Recognition: Special Issue on Grammatical Inference
  Techniques \& Applications}, 38/9:1349--1371, 2005.

\bibitem[DE03]{DenisEsposito2003}
F.~Denis and Y.~Esposito.
\newblock Residual languages and probabilistic automata.
\newblock In {\em 30th International Colloquium, ICALP 2003}, number 2719 in
  LNCS, pages 452--463. SV, 2003.

\bibitem[DE04]{DenisEsposito04}
F.~Denis and Y.~Esposito.
\newblock Learning classes of probabilistic automata.
\newblock In {\em COLT 2004}, number 3120 in LNAI, pages 124--139, 2004.

\bibitem[DEH06]{DenisEspositoHabrardTR06}
F.~Denis, Y.~Esposito, and A.~Habrard.
\newblock Learning rational stochastic languages.
\newblock Technical Report ccsd-00019161, HAL, 2006.
\newblock https://hal.ccsd.cnrs.fr/ccsd-00019161.

\bibitem[dlHT00]{HigueraThollard00}
C.~de~la Higuera and F.~Thollard.
\newblock Identification in the limit with probability one of stochastic
  deterministic finite automata.
\newblock In {\em Proceedings of the 5th ICGI}, volume 1891 of {\em LNAI},
  pages 141--156. Springer, 2000.

\bibitem[DLR77]{DempsterLairdRubin77}
A.P Dempster, N.~M. Laird, and D.~B. Rubin.
\newblock Maximum likelihood from incomplete data via the em algorithm.
\newblock {\em Journal of the Royal Statistical Society}, 39:1--38, 1977.

\bibitem[DLT02]{DenisLemayTerlutte2002b}
F.~Denis, A.~Lemay, and A.~Terlutte.
\newblock Residual {F}inite {S}tate {A}utomata.
\newblock {\em Fundamenta Informaticae}, 51(4):339--368, 2002.

\bibitem[DLT04]{DenisLemayTerlutte04}
F.~Denis, A.~Lemay, and A.~Terlutte.
\newblock Learning regular languages using rfsas.
\newblock {\em Theoretical Computer Science}, 2(313):267--294, 2004.

\bibitem[ELDD02]{EspositoLemayDenisDupont2002}
Y.~Esposito, A.~Lemay, F.~Denis, and P.~Dupont.
\newblock Learning probabilistic residual finite state automata.
\newblock In {\em ICGI'2002, 6th ICGI}, LNAI. Springer Verlag, 2002.

\bibitem[Fli74]{Fliess74}
M.~Fliess.
\newblock Matrices de {H}ankel.
\newblock {\em J. Maths. Pures Appl.}, 53:197--222, 1974.
\newblock + erratum in Vol. 54 (1975).

\bibitem[Gan66]{Gantmacher66}
F.~R. Gantmacher.
\newblock {\em Th\'eorie des matrices, tomes 1 et 2}.
\newblock Dunod, 1966.

\bibitem[Jac75]{Jacob75}
G.~Jacob.
\newblock Sur un th\'eor\`eme de {S}hamir.
\newblock {\em Information and control}, 1975.

\bibitem[Sak03]{Sakarovitch03}
Jacques Sakarovitch.
\newblock {\em \'El\'ements de th\'eorie des automates}.
\newblock \'Editions Vuibert, 2003.

\bibitem[Sch61]{Schutzenberger61}
M.~P. Sch\"utzenberger.
\newblock On the definition of a family of automata.
\newblock {\em Information and Control}, 4:245--270, 1961.

\bibitem[SS78]{SalomaaSoittola78}
Arto Salomaa and M.~Soittola.
\newblock {\em Automata: Theoretic Aspects of Formal Power Series}.
\newblock Springer-Verlag, 1978.

\end{thebibliography}
